%% file: main.tex
\documentclass[runningheads]{llncs}

\usepackage{eccv}

\usepackage{eccvabbrv}

\usepackage{graphicx}
\usepackage{booktabs}
\usepackage{multirow}
\usepackage[accsupp]{axessibility}  
\newcommand{\acc}{$Acc_{s \rightarrow t}$}

\usepackage{comment}
\usepackage{wrapfig}

\usepackage{color, colortbl}
\definecolor{Ours}{rgb}{0.9,1,0.9} 
\definecolor{Gray}{gray}{0.95}
\usepackage{hhline}

\usepackage[pagebackref,breaklinks,colorlinks]{hyperref}

\usepackage{orcidlink}

\begin{document}

\title{On Inherent Adversarial Robustness of Active Vision Systems} 

\titlerunning{RAVS}

\author{Amitangshu Mukherjee\inst{1}\orcidlink{0000-0003-2704-3580} \and
Timur Ibrayev\inst{1}\orcidlink{0000-0001-8849-8971} \and
Kaushik Roy\inst{1} \orcidlink{0000-0002-0735-9695}}

\authorrunning{A.~Mukherjee et al.}

\institute{Purdue University, West Lafayette IN 47906, USA 
}

\maketitle

\begin{abstract}
Current Deep Neural Networks are vulnerable to adversarial examples, which alter their predictions by adding carefully crafted noise. Since human eyes are robust to such inputs, it is possible that the vulnerability stems from the standard way of processing inputs in one shot by processing every pixel with the same importance. In contrast, neuroscience suggests that the human vision system can differentiate salient features by $(1)$ switching between multiple fixation points (saccades) and $(2)$ processing the surrounding with a non-uniform external resolution (foveation). In this work, we advocate that the integration of such active vision mechanisms into current deep learning systems can offer robustness benefits. Specifically, we empirically demonstrate the inherent robustness of two active vision methods—GFNet and FALcon—under a black box threat model. By learning and inferencing based on downsampled glimpses obtained from multiple distinct fixation points within an input, we show that these active methods achieve \textbf{2-3} times greater robustness compared to a standard passive convolutional network under state-of-the-art adversarial attacks. More importantly, we provide illustrative and interpretable visualization analysis that demonstrates how performing inference from distinct fixation points makes active vision methods less vulnerable to malicious inputs.

\end{abstract}

\section{Introduction}
\label{sec:intro}

\input{Introductions/proposed_introduction}


\section{Related Work}
\label{sec:related}

In this section, we highlight key related works, while a comprehensive study of related literature is provided in the Supplementary.

\textbf{Active Vision Methods} The methods discussed here explore the incorporation of active iterative strategies for input processing. RANet \cite{Mnih2014RecurrentMO} incorporates a recurrent attention network to selectively focus on different parts of the input sequence over multiple time steps excelling in sequential tasks. Saccader \cite{Saccader} emulates saccadic eye movements to iteratively extract features from an image attending to finer details while enhancing performance. Glance and Focus Networks (GFNet) \cite{GFNet_2020} constrained by computational budget, iteratively processes different glimpses in an image, refining predictions until confidently identifying the object. Foveated Transformer \cite{jonnalagadda2022foveater}  uses pooling regions and dynamic fixation allocation based on Transformer attention based on past and present fixations for image classification. FALcon \cite{ibrayev2023exploring} employs an active vision strategy to learn from multiple distinct fixation points, effectively capturing entire objects. In this study, we explore how the iterative interplay of foveation and saccades enhances the inherent adversarial robustness of active methods in a black-box setting.

\textbf{Towards Bio-Inspired mechanism for Robustness} 
The following methods address the adversarial inputs by functionally treating human eyes as pre-processing/transformation stage.
R-Warp \cite{Poggiocortical} advocates for biologically inspired mechanisms such as cortical fixations and retinal fixations incorporated in DNNs lead to adversarial robustness for small perturbations. VOneBlock \cite{VOneBlock}, illustrates that incorporating primary visual cortex processing at the forefront of CNNs enhances their resilience against image perturbations. Harrington et al. \cite{harrington2022finding} demonstrates that adversarially robust networks behave similarly to texture peripheral vision models, thus promoting the latter's plausibility for adversarial robustness. Gant et al. \cite{gant2021evaluating} proposed a novel Foveated Texture Transform module in a VGG-11 to enhance adversarial robustness without sacrificing standard accuracy. R-Blur \cite{shah2023training} simulates peripheral vision using adaptive Gaussian blurring and trains on these transformed input images, leading to improved adversarial robustness. While these methods simulate human peripheral processing, they do not replicate the iterative active learning process found in human vision. We provide a fresh perspective showing how mimicking human-like active vision processing naturally enhances DNN robustness against adversarial inputs.


\textbf{Transfer Attacks} Szegedy et al. \cite{szegedy2014intriguing} introduced the vulnerabilities of neural networks to adversarial samples.  Papernot et al. \cite{Papernot2016PracticalBA} introduced a novel approach that leverages substitute models to craft transferable adversarial examples, emphasizing the need for robust defenses against such attacks. Liu et al. \cite{liu2017delving} conducted an extensive investigation into the transferability of adversarial samples on large-scale datasets like ImageNet \cite{ImageNet}. Recently, LGV \cite{LGV_eccv22} exploited the weight space geometry of surrogate models to find flatter adversarial samples creating stronger transfer attacks. In this study, we examine active methods such as GFNet and FALcon under the lens of adversarial robustness in a black box threat model and show the human-inspired active way of processing inputs in DNNs leads to inherent robustness.


\section{Active Vision Methods}
\label{sec:avm}

In this section, we provide a focused overview of the inference process and highlight key insights into the inherent robustness of two active vision methods: Glance and Focus Networks (GFNet) \cite{GFNet_2020}, and FALcon \cite{ibrayev2023exploring}. These methods simulate foveation by cropping glimpses from the image based on fixation points, without blurring the extracted glimpses. This can be interpreted as foveation with an extreme cut-off. For learning details, readers are directed to Supplementary Sections 1.2 (GFNet) and 1.3 (FALcon).  

\subsection{Glance and Focus Networks}
\label{ssec:gfinf}


\begin{figure}[tb]
  \centering
  \includegraphics[height=6.0 cm]{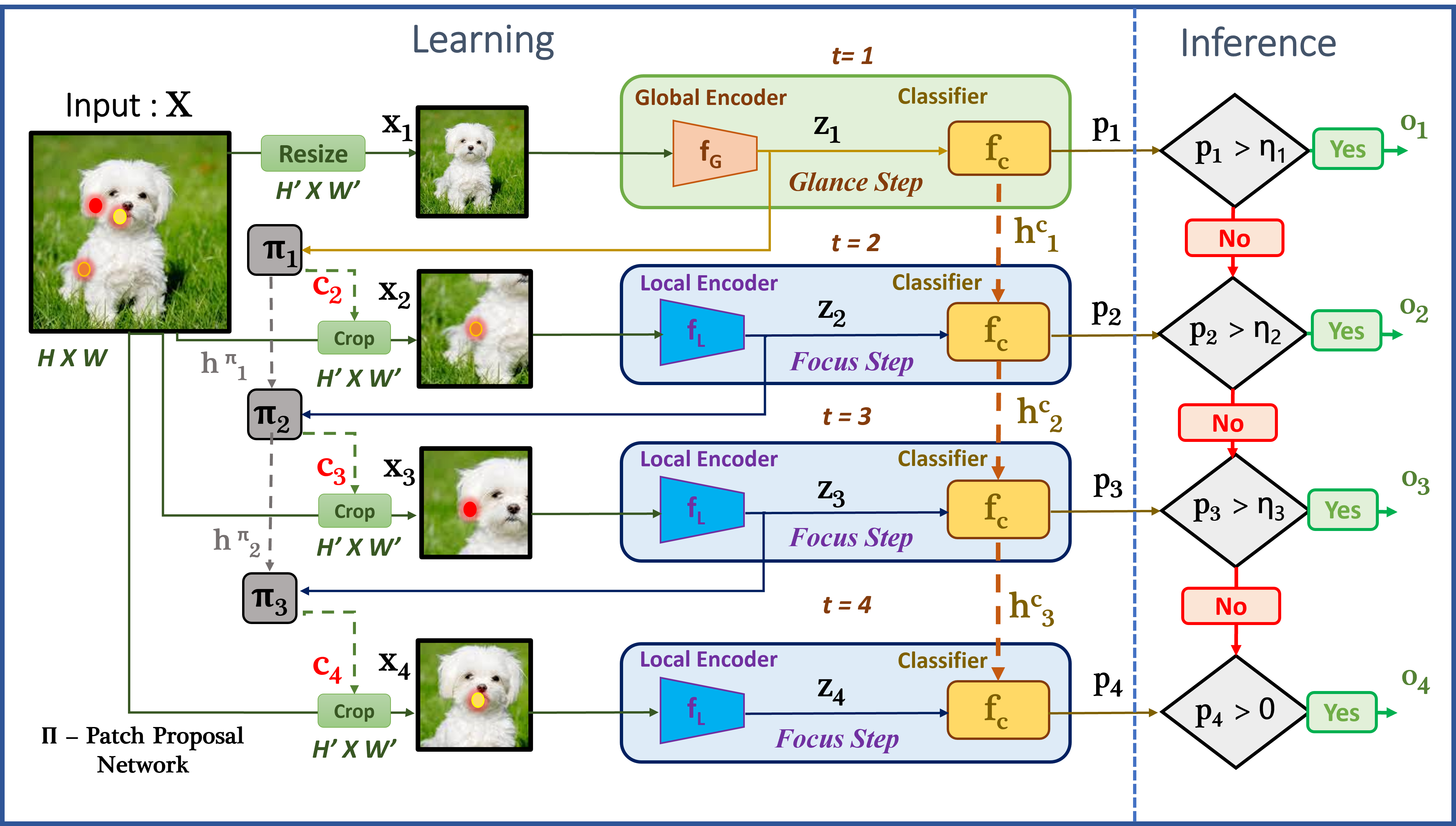}
  \caption{(\textbf{Learning \& Inference}) The figure provides an overview of GFNet's operation. It begins by downsampling the input image to a lower resolution for rapid prediction $(p_{1})$, termed \textbf{$f_{G}$ (Glance)} at $t=1$. If the network lacks confidence $(p_{1} < \eta_{1})$, it enters subsequent \textbf{$f_{L}$ (Focus)} steps until certainty is attained or till $(t=4)$. Each focus step analyzes a patch $(H' \times W')$ cropped  from the original input $(H \times W)$ centered around $(c_{t})$ illustrated by colored dots. These co-ordinates are determined by a patch proposal network \textbf{$\pi$}. The process is depicted for a sequence length of $4$. 
  \vspace{-5mm}}
  \label{fig:gfblock}
\end{figure}


GFNet performs inference in two distinct steps -- a glance step and subsequent multiple focus steps (Figure~\ref{fig:gfblock}). In the glance step, the full-resolution image $(H \times W)$ is first downsampled to a much lower resolution $(H' \times W')$. Then it is passed through a global encoder $f_{G}$ to make a swift prediction based on the global features.  If the confidence $p_{t}$ exceeds the threshold $\eta_{t}$, where $\eta$ is a pre-defined threshold 
\cite{huang2018multiscale, yang2020resolution}, the process halts. A patch-proposal network evaluates these features concurrently to predict the fixation point for the subsequent focus step. Each patch is generated based on the most salient features of the object centered around the fixation point. This is denoted by the orange dot in the second patch. This $H' \times W'$ patch is cropped from the image and inputted into the local encoder $f_{L}$. Simultaneously coordinates for the next focus step are produced by the patch-proposal network. The iterative process persists until the network gains sufficient confidence in its prediction, or reaches the end of the sequential process $t = T$. To understand the robustness aspect through inference on transferred adversarial samples, we keep the early termination inactive, allowing inference to continue until $t = T$. This provides the network distinct fixation point, at each step to assess the input in the presence of adversarial noise. The global and local encoders $f_{G}$ and $f_{L}$ respectively are deep convolutional neural networks and are trained on the low dimensional inputs of $H' \times W'$. Exploiting this in our experiments, we demonstrate how learning in a downsampled resolution contributes to the robustness properties of such methods \ref{ssec:glimpses}.

\subsection{FALcon}
\label{ssec:falinf}

\begin{figure}[tb]
  \centering
  \includegraphics[height=6.00 cm]{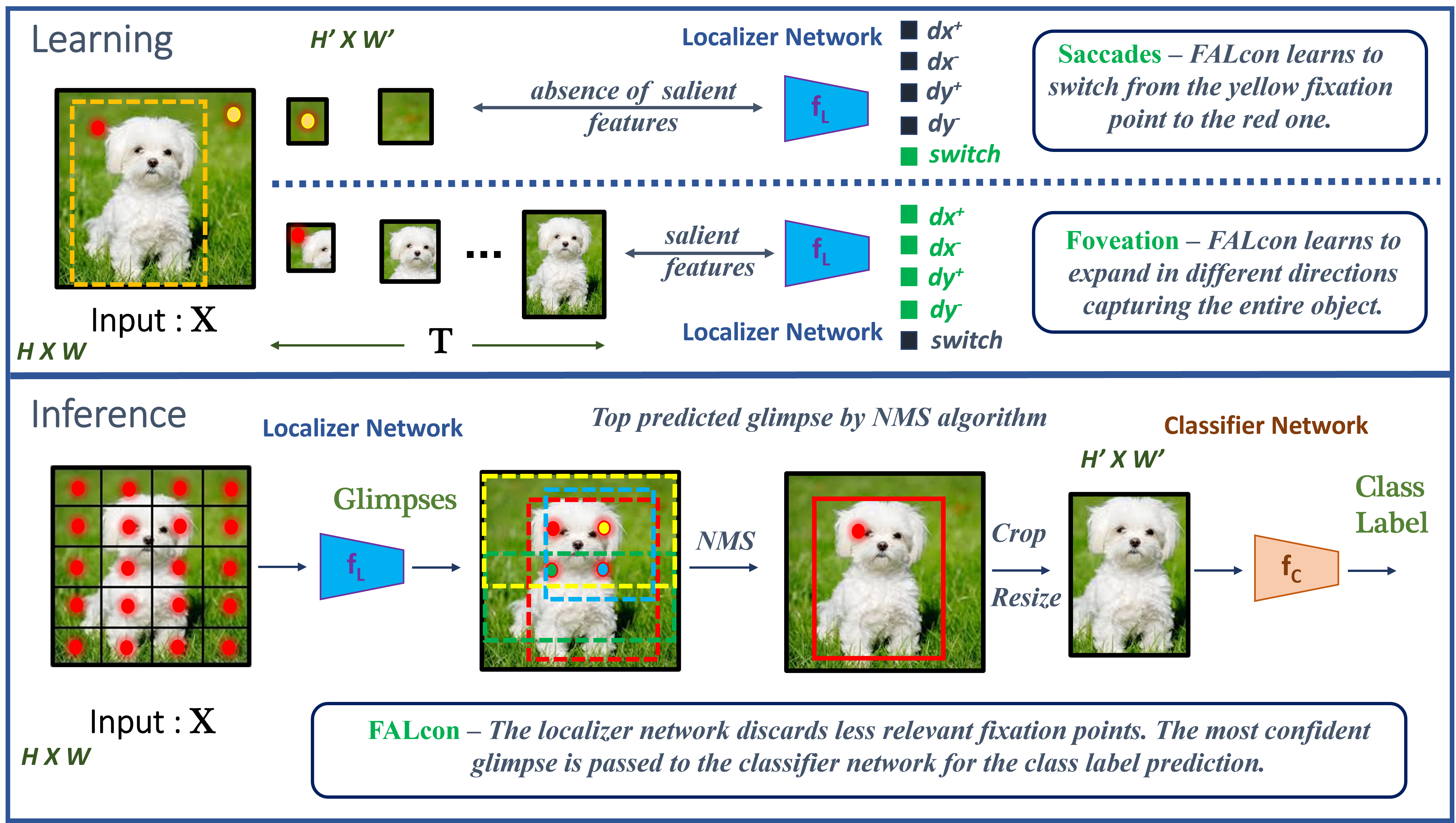}
  \caption{The figure provides a high-level overview of FALcon. During \textbf{Learning}, a Localizer network $(f_{L})$ is trained to predict five distinct actions (four for expansion and one for switching), enabling it to learn the importance of each fixation point illustrated by colored dots. Learning occurs in a downsampled resolution of $(H' \times W')$. During \textbf{Inference}, $f_{L}$ starts from each pre-defined multiple fixation point (20 red dots). If salient object features are present, $f_{L}$ performs the learned expansions to capture the object (4 colored dashed boxes, colored dots). The most confident final foveated glimpse (red solid box) is cropped $(H' \times W')$ and presented to the classifier $(f_{C})$. 
  \vspace{-5mm}}
  \label{fig:falcon}
\end{figure}




During inference, the input image $X$ is divided into grid cells, as illustrated in the first image at the bottom part of Figure~\ref{fig:falcon}. Each grid cell is considered as an initial fixation point (red dot). The localizer $f_{L}$ initiates from each point with a pre-defined glimpse size to inspect salient object features. In the absence of salient features, the localizer deems there is no potential object and hence switches to another fixation point. If the localizer encounters any relevant features and hence assumes the presence of a potential object, it expands from the initial glimpse dimensions in four independent directions to generate a sequence of foveated glimpses. Based on the expansion operations, each foveated glimpse is cropped from the input image at a downsampled resolution of $H' \times W'$. The sequence of expanding for a potential fixation point continues until the final foveated glimpse captures the entire object. This process iterates for all potential points that could lead to various final foveated glimpses for the same object. This is illustrated by the colored dots and dashed colored boxes in the second image. For object localization, these final foveated glimpses undergo non-maximum suppression to yield the most confident prediction indicated by the solid red bounding box in the third image. The region corresponding to the most confident final foveated glimpse is cropped at dimensions $H' \times W'$, and then presented to the classification network $f_{C}$ for class prediction. In our evaluation, we only use this final class prediction label. The remaining potential dots, except the most confident one although not utilized for localization, serve as a map to understand the effect of non-uniform adversarial noise injected into images. In our experiments, we visually demonstrate how the noise affects some of these fixation points, but not all (\ref{ssec:saccades}) resulting in enhanced robustness. Both localizer $f_{L}$ and classifier $f_{C}$ are deep convolutional neural networks. During learning, $f_{C}$ is not fine-tuned on the lower dimensional inputs as is done for GFNets.



\section{Methodology}
\label{sec:method}

In this section, we introduce the general setup for our experiments outlined in Section~\ref{sec:exps}. We follow the adversarial transfer attack protocol for a black box threat model as outlined in \cite{liu2017delving, AdvVit}. Following this protocol we define non-targeted transferability. Given a surrogate Classifier $S_{i}$, we generate an adversarial sample for an image/label pair $(x,y)$ which is denoted as $x_{adv}$. This is with respect to the surrogate Classifier $S_{i}$ and attack pair $A_{S_{i}}$. Now this adversarial sample $x_{adv}$ is said to transfer to another target Classifier $T_{i}$ if the adversarial sample is mispredicted. This is formalized as the following:


\begin{equation}
\begin{aligned}
x_{adv} = A_{S_{i}}(x,y) \mid S_{i}(x_{adv}) \neq y ; &\qquad\qquad 
    T_{i}(x_{adv}) \neq y    
\end{aligned} 
\end{equation}


\textbf{Metrics} We measure the non-targeted transferability by computing the percentage of adversarial examples generated using model $S_{i}$, but still correctly classified
by the model $T_{i}$ (not transferred). We refer to this percentage as accuracy. A higher accuracy means less susceptibility to transferred adversarial samples and hence higher robustness. For a test set with $N$ samples, the accuracy is defined as \acc: 

\begin{equation}
    {1/N}\sum_{j=1}^{N} T_{i}({x_{adv}}_{j}) = y_{j}      
\end{equation}


\textbf{{Remark}} In this study, we focus solely on empirically showcasing the inherent robustness of active vision methods. We do not propose any adversarial defense for a black-box attack scenario or analyze the transferability trends between surrogates and target samples. Therefore, we opt for standard accuracy (\acc).


\section{Experiments}
\label{sec:exps}

The experimental section aims to illustrate that active vision networks GFNet \cite{GFNet_2020} and FALcon \cite{ibrayev2023exploring} exhibit higher levels of robustness against transferred adversarial images (\ref{sec:method}) than base passive classifiers \cite{He2015DeepRL}. Section~(\ref{ssec:blackbox}) empirically demonstrates this via quantitative results. Sections~(\ref{ssec:glimpses}) and (\ref{ssec:saccades}) then focus on explaining the origins of inherent robustness by analysing the internal mechanics of GFNet and FALcon, respectively, in the presence of adversarial inputs.

\subsection{Implementation Details}
\label{ssec:impl}

We perform our extensive robustness analysis on Imagenet \cite{ImageNet}, a standard benchmark for image classification. We utilize ImageNet pre-trained weights for GFNet and FALcon without any additional fine-tuning. We employ GFNets with Res-Net50 as global \(f_{G}\) and local \(f_{L}\) encoders, as illustrated in Figure~\ref{fig:gfblock}. Both encoders are trained on downsampled resolution images of (96, 96). For FALcon we employ VGG16 \cite{simonyan2014very} as localizer \(f_{L}\), while ResNet50 as classifier \(f_{C}\). Please note that, unlike GFNet, the \(f_{C}\) of FALcon is not fine-tuned on downsampled images during training. Instead, only the localizer is trained on the downsampled images. We utilize \textsc{Torchattacks} \cite{kim2020torchattacks}, an integrated library for generating adversarial attacks \cite{ravikumar2022trend} with \textsc{PyTorch}, to generate adversarial samples.  


\subsection{Inherent Robustness in the Black box Setup}
\label{ssec:blackbox}
In this section, we demonstrate the enhanced performance of active vision methods (e.g., FALcon and GFNet) compared to passive baseline methods (e.g., supervised-trained ResNet) in a black-box setup. For GFNet, we utilize the output at the final step of the prediction sequence, as described in Section~(\ref{ssec:gfinf}). For FALcon, we utilize the most confident class prediction as detailed in Section~(\ref{ssec:falinf}). All target classifiers $(T_{i})$ share the same common backbone architecture, which is ResNet50. Similarly, we consider an adversarially trained ResNet50 \cite{madry2018towards}, which serves as an Oracle method denoted as Adv-T$\star$. Trained specifically for adversarial defense, this method offers the best-case performance on transferred samples. The comparison of interest is however between the active methods and the passive baselines. We want to demonstrate the inherent robustness of active methods that do not necessitate additional targeted training for adversarial defense. However, observing Adv-T$\star$ results, one might look into an opportunity for adapting adversarial training methods into active vision methods.

\input{Tables/InherentRobustness}



\textbf{Iterative Attacks} We generate adversarial samples from surrogate classifiers \(S_{i}\) using iterative adversarial attacks such as PGD \cite{madry2018towards}, MIFGSM \cite{MIM}, VMIFGSM \cite{VMI} and Patchwise-IFGSM \cite{GaoZhang2020PatchWise}. We conduct \( L_{\infty} \) attacks with 10 iterative steps, \( \alpha = 2/255 \), and \( \epsilon = 8/255 \) for all four iterative attacks. Adversarial samples are generated on the entire test set of the ImageNet dataset of 50,000 samples. The corresponding clean accuracy is shown at the top of Table (\ref{tab:Main}).  

The results shown in Table~(\ref{tab:Main}) demonstrate the superior performance of both the active vision methods over the passive approach across all surrogate architectures and attacks. All $T_{i}$ except Adv-T$\star$, exhibit performance degradation proportional to the strengths of the attacks. Similarly, except for the Adv-T$\star$, all $T_{i}$ experience a performance drop when the surrogate network $S_{i}$ shares the architecture of the same classification backbone. Nevertheless, FALcon and GFNet, due to their respective processing mechanisms, provide an additional shield even when the attack is generated based of the shared backbone architecture. 



\textbf{Transfer Attacks with Large Geometric Vicinity (LGV)} Table~(\ref{tab:LGV_table}) presents results based on a geometric space attack \cite{LGV_eccv22}. We follow a similar experimental setup as outlined in the paper \cite{LGV_eccv22}, combining LGV with PGD and BIM \cite{kurakin2018adversarial} on 1000 randomly sampled images from the ImageNet validation set. We report accuracy (\acc), and the results indicate a consistent trend. We also incorporate with CutMix \cite{CutMix} into the analysis as an additional passive method, because of its known robustness properties stemming from its strong regularized feature representations. As expected, it shows more robustness than the baseline passive method but still exhibits inferior performance compared to the active methods. In the Supplementary, we offer additional results utilizing various surrogate classifiers. Additionally, we provide a performance analysis on Token Gradient Regularization \cite{zhang2023transferable}, a SOTA transfer attack based on transformers that has shown to be effective against CNNs as well.

In the following sections, we present the primary factors contributing to this enhanced robustness: processing inputs in a down-sampled resolution (Section~\ref{ssec:glimpses}) and performing inference from different fixation points (Section~\ref{ssec:saccades}).


\input{Tables/Horizontal_Geometric_Attack_Token_Grad}

\subsection{Effects of glimpse-based downsampling (case study: GFNet)}
\label{ssec:glimpses}

In this section, we use GFNet to explore how learning image representations based on glimpses at a downsampled resolution contributes to the inherent robustness. Downsampling inherently causes reduction in features. Adversarial imperceptible noise is crafted based on the image in its original resolution (e.g. 224 $\times$ 224). Hence downsampling the image, distorts the noise along with it, thereby reducing its overall impact on predictions. As a result, it is probable to think that an inherent robustness offered by models processing an image via downsampled resolution stems from the distortions on the non-uniform adversarial noise. To analyse this factor we organize experiments in this section into 3 settings:


\input{Tables/GlimpseDown}

\begin{itemize}
    \item \textbf{Setting $1$} \textit{Effect of processing downsampled clean images} - Images from the test set are used for evaluation without any adversarial attack. The images are downsampled to $(96,96)$ and $(128,128)$ and inference is performed.    
    \item \textbf{Setting $2$} \textit{Reduction of efficacy of adversarial noise post downsampling} - Adversarial images are first generated from full resolution images of $(224,224)$ and then downsampled to $(96,96)$ and $(128,128)$, separately, for inference. for an active vision method such as the GFNet, this is an inherent step of their learning and inference pipeline. However, for passive vision methods, we resize the adversarial inputs to match the resolutions separately.   
    \item \textbf{Setting $3$} \textit{Generating adversarial attacks on downsampled images} - The images are downsampled to lower resolutions first and then adversarial inputs are generated. These adversarial downsampled inputs are then passed for inference on both passive and active target models. Since downsampling is performed first, the adversarial effect is not downgraded.
\end{itemize}


For the passive target baseline, we use a ResNet50 pre-trained on ImageNet at resolutions of $224 \times 224$. For GFNets, we infer with two separate models trained on $96 \times 96$ and $128 \times 128$ resolutions. Notably, we maintain consistency by evaluating GFNets on images of matching resolutions. To illustrate downsampling effects, passive baselines are tested on downsampled images of $96$ and $128$ resolutions (see Table~\ref{tab:glimpse_down}). For simplicity, we further refer to GFNets trained on $96 \times 96$ dimensions as "GFNet-96".

\textbf{Results} Table~\ref{tab:glimpse_down} presents quantitative results, focusing on \acc. The best performing models are highlighted in bold. For Setting $1$, a passive model trained on a higher resolution suffers a drop in performance when evaluated at downsampled input, unlike GFNets trained for downsampled resolutions. Setting $2$ shows that simply downsampling adversarial images to lower resolutions is beneficial. This indicates along with the image resolution, the imperceptible adversarial noise also probably gets downsampled thereby reducing its effect on model predictions even when $T_{i}$ is same as $S_{i}$. Furthermore, under this setting, GFNet-96 exhibits greater inherent robustness than GFNet-128 when compared to their corresponding passive baselines. For Setting $3$, it is evident that all target models suffer a drop in performances indicating that generating adversarial inputs at resolutions corresponding to target models leads to more potent attacks. Remarkably, GFNet-96 and GFNet-128 demonstrate performance improvements close to \textbf{$3 \times$} and \textbf{$2 \times$}, respectively for ResNet34 as $S_{i}$, compared to their corresponding passive baselines on downsampled adversarial samples. This further emphasizes the effectiveness of learning in a downsampled regime even under the presence of adversarial attacks.      

\subsection{Effect of distinct fixation points (case study: FALcon)}
\label{ssec:saccades}

\begin{figure}[tb]
  \centering
  \includegraphics[height=4.3 cm]{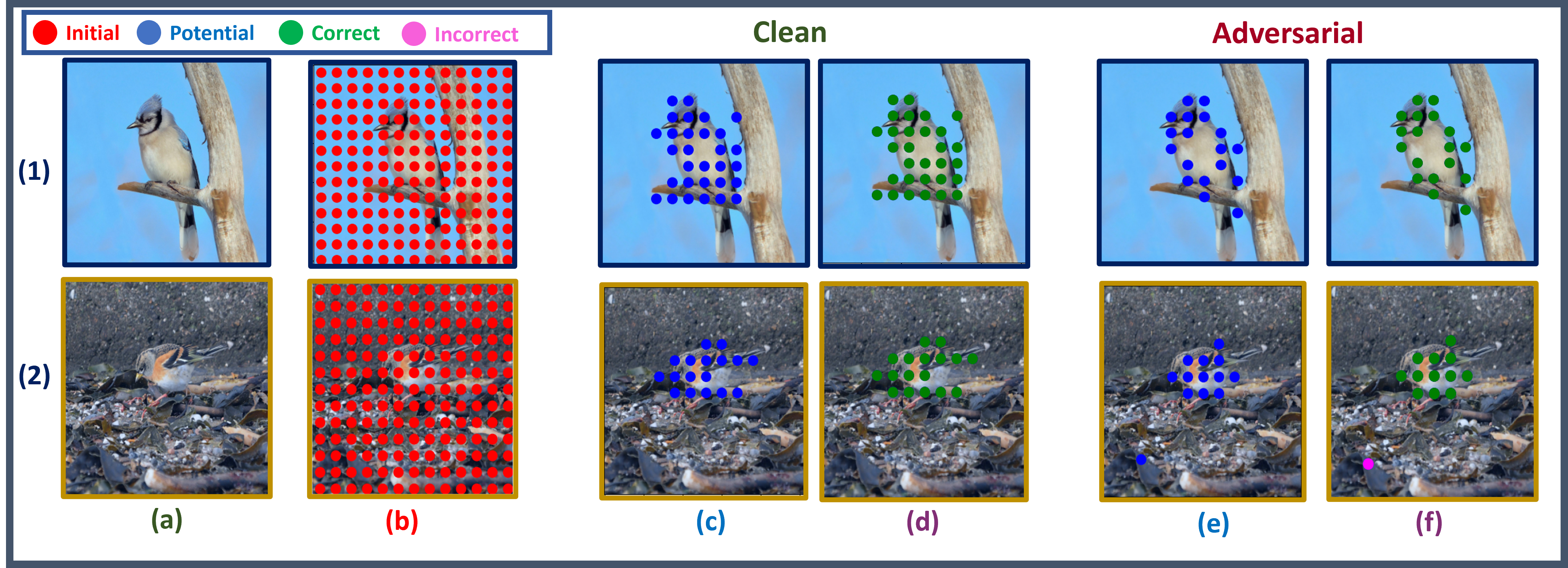}
  \caption{Figure illustrates Initial Fixation Point Maps \textbf{(IFPM)} to show the efficacy of performing inference from multiple fixation points. An \textbf{IFPM} is a visual representation that depicts the spatial locations of the initial starting positions of FALcon. \textbf{(b)} illustrates all \textcolor{red}{initial fixations} points via gridding for both clean and adversarial inputs. \textbf{(\textcolor{blue}{c} \& \textcolor{Plum}{d})} show the \textcolor{blue}{potential} \textbf{and} \textcolor{Plum}{evaluated} initial fixation points for a clean sample. Similarly, \textbf{(\textcolor{blue}{e} \& \textcolor{Plum}{f})} show the same for an adversarial sample. An \textcolor{Plum}{evaluated} \textbf{IFPM} can consist of both \textcolor{OliveGreen}{correct} and \textcolor{Magenta}{incorrect} points as denoted by \textcolor{Plum}{2f}. Adversarial noise spreads non-uniformly across an image and affects different initial points differently. This is indicated by the reduced number of \textcolor{blue}{potential} \textbf{(\textcolor{blue}{c} to \textcolor{blue}{e})} and \textcolor{OliveGreen}{correct} points \textbf{(\textcolor{Plum}{d} to \textcolor{Plum}{f})} from a clean \textbf{to} an adversarial sample. Still, the presence of a positive number of \textcolor{OliveGreen}{correct points} \textbf{(\textcolor{Plum}{f})} underscores the inherent robustness of an active method. 
  \vspace{-5mm}
  }
  
  \label{fig:FPM_image}
\end{figure}

In this section, we use FALcon to demonstrate the effect of processing an image from distinct fixation points on the robustness of active vision methods. The capability of FALcon to consider various fixation points is used to extract interpretable visualization results. Moreover, since FALcon has independent models for localization and classification, with the latter model not fine-tuned during training, it allows for a fair comparison with passive baseline network.

\subsubsection{Initial Fixation Point Map}


In order to understand the impact of adversarial noise on regions of the image that influence model predictions, we define an Initial Fixation Point Map (IFPM). IFPM displays the distribution of initial fixation points based on how each of them affects the decision-making of FALcon throughout the inference process. Figure~\ref{fig:FPM_image} shows IFPMs generated for both clean and adversarial images. As described in Section~\ref{ssec:falinf}, FALcon processes every input from multiple initial fixation points. \textcolor{red}{Red} dots indicate all initial fixation points, equally distributed over the image dimensions. Each point is then presented to the localizer, which retains only those, indicated by \textcolor{blue}{blue} dots, that potentially resulted in the capture of an object through the series of expanding foveated glimpses. 
The classifier processes the final foveated glimpses that resulted from potential points to determine the class label of an object. As a result, various fixation points result in FALcon making correct or incorrect output predictions, indicated by \textcolor{OliveGreen}{green} and \textcolor{magenta}{magenta} dots, respectively. By obtaining IFPM for clean and adversarial versions of the same image, we illustrate how the adversarial noise impacts FALcon in terms of its capacity to make correct predictions from various fixation points.

\begin{wrapfigure}{R}{0.5\textwidth}
  \begin{center}
    \includegraphics[width=0.48\textwidth]{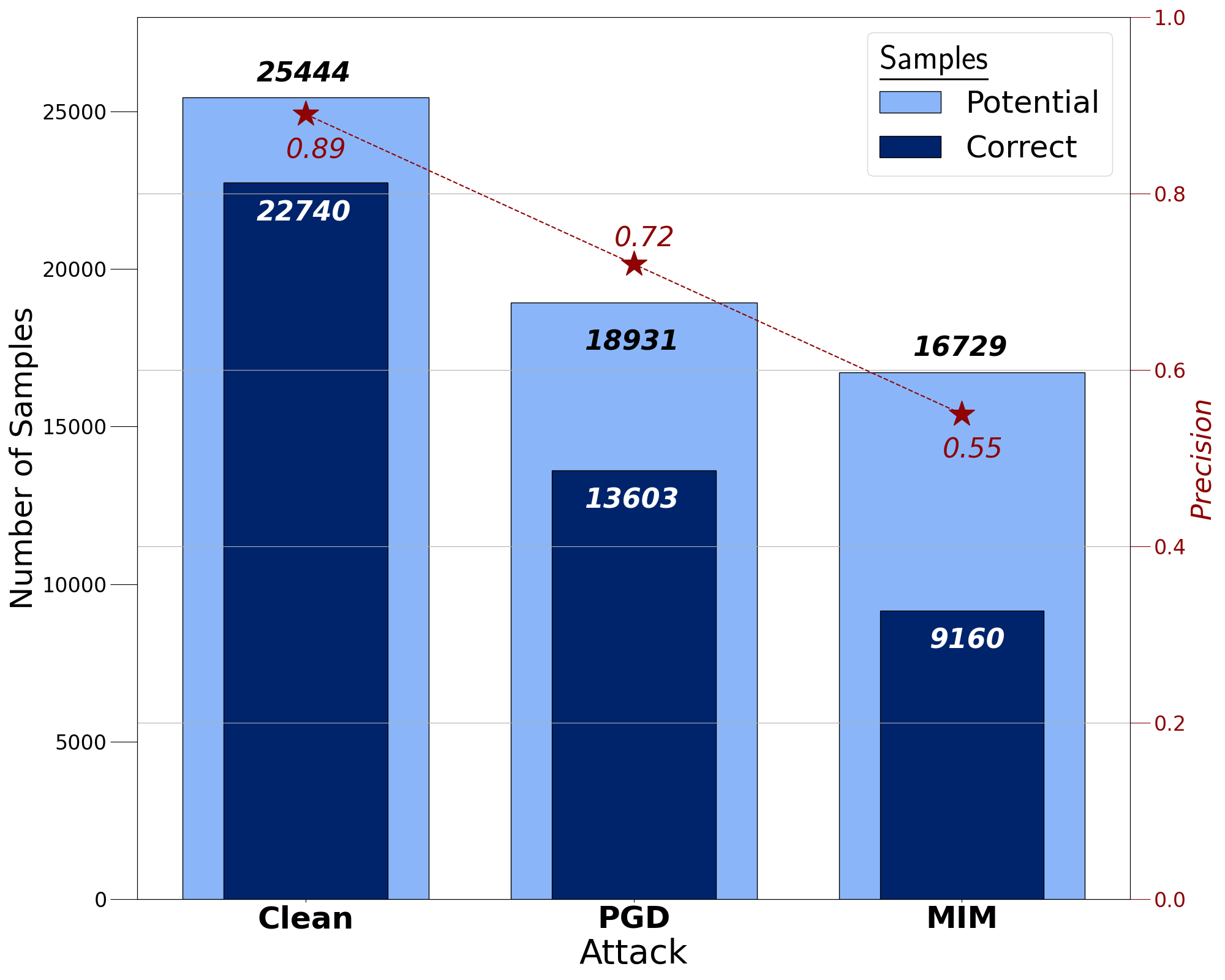}
  \end{center}
  \vspace{-6.5mm}
  \caption{Precision of predictions}
  \label{fig:Fixation_PM_tab}
  \vspace{-5mm}
\end{wrapfigure}

\textbf{Results} IFPMs illustrated in Figure~\ref{fig:FPM_image} show that despite the addition of adversarial noise, multiple initial fixation points result in correct final predictions (d \& f). IFPM clearly indicates the reduced number of potential and correct points for an adversarial sample compared to the corresponding clean sample (c to e) and (d to f). Due to its non-uniformity and imperceptible criteria, the adversarial noise does not affect each point equally. Hence, multiple fixation points lead to correct class predictions. This visually explains the reason for the improved performance of an active method over a passive one, supporting the quantitative results presented in the previous sections. Although noise affects the method, its inherent processing from multiple fixations makes it less susceptible (f). In the second sample (2f), we can notice of a \textcolor{magenta}{magenta} fixation point far away from the object. This is not present for the clean sample and is a false positive due to the addition of the noise. Yet, around the object, we can see multiple green points indicating correct prediction. This validates the hypothesis presented in the third column in Figure (\ref{fig:block}).

Apart from \acc, precision defined as $correct/potential$ can be considered as another metric to account for the enhanced robustness. Potential points can be viewed as the sum of true positive (green) and false positive (magenta) as illustrated by an IFPM.  We generate attacks for 1000 images using ResNet34 as the surrogate model. As can be seen from Figure~\ref{fig:Fixation_PM_tab}, the high precision for clean samples, decreases for adversarial samples depending on the strength of the respective attack. Still, the persistence of a high number of true positives, provides a quantitative justification for the improved performance of FALcon on adversarial samples.

\subsubsection{Explaining Adversarial Vulnerability of Passive Methods}
\label{sssec:viz}

\begin{figure}[tb]
  \centering
  \includegraphics[height=6.5 cm]{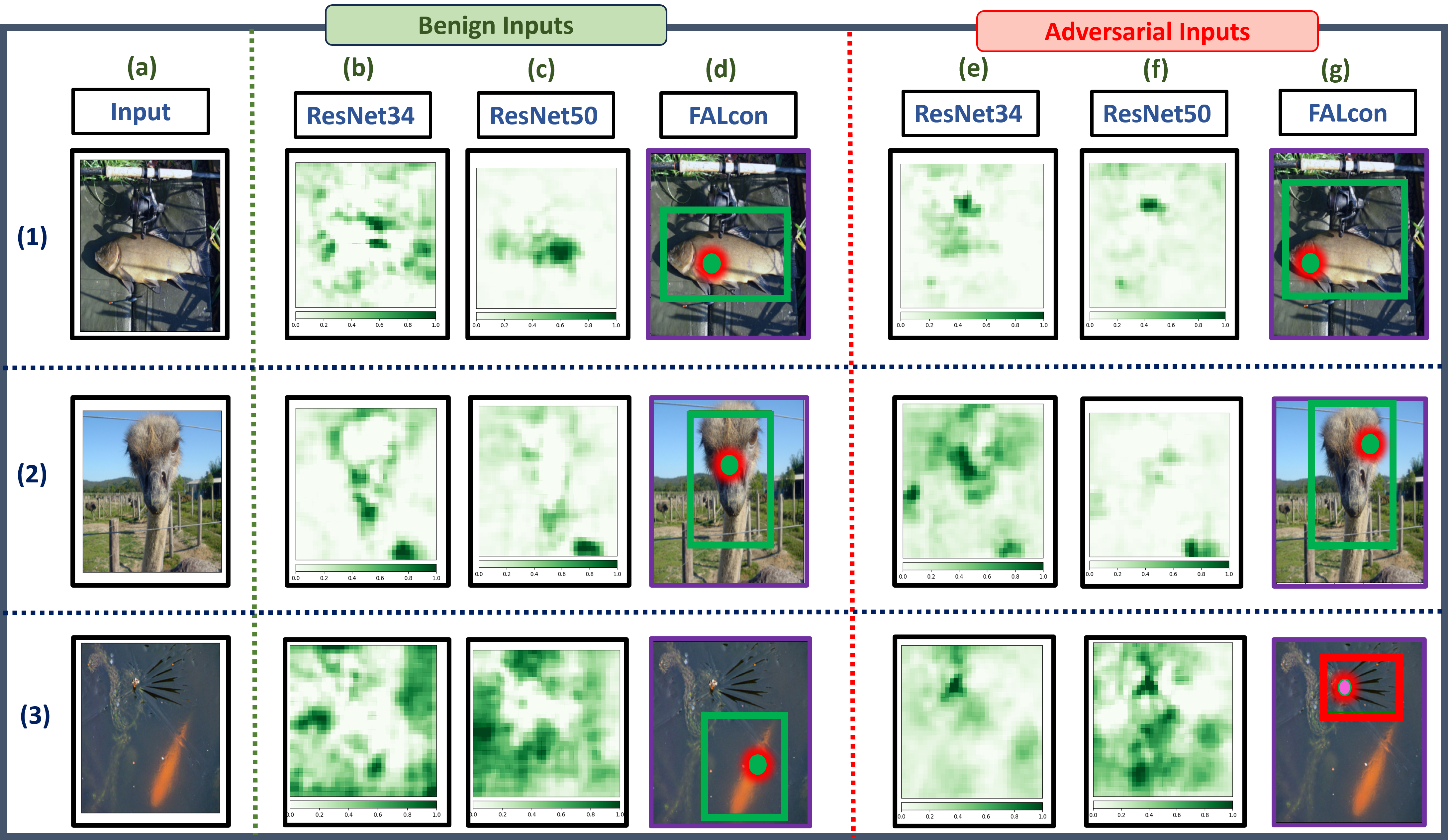}
  \caption{The figure illustrates vulnerabilities of passive methods to adversarial samples. 
  For passive methods \textbf{(b,c,e,f)}, occlusion maps are provided, highlighting areas responsible for model predictions. For FALcon \textbf{(d,g)}, the final foveated glimpses along with the corresponding initial fixation points are presented. Even though adversarial noise affects an image non-uniformly, passive methods struggle to evade the noise as they process the entire image with equal importance. Contrary, the final foveated glimpse highlights the effect of adversarial noise, guided by the corresponding fixation point.     \vspace{-4mm}
  }
  \label{fig:Occ_image}
\end{figure}

As mentioned earlier, the probable cause of adversarial vulnerability of the passive vision methods is that they process an input in one-shot with uniform resolution, where every input pixel is treated with the same importance. This is visually demonstrated in this section via occlusion maps. Figure~\ref{fig:Occ_image} illustrates occlusion maps for passive methods (b,c,e,f) and the final foveated glimpse for FALcon (d,g). An occlusion map is a visual heatmap indicating key regions of an image when occluded, affect the model performance. The darker the region, the higher contribution it has on the final prediction. Occlusion maps are generated based on prediction labels. We first generate the adversarial sample and then generate the occlusion map based on the predicted adversarial label. We use ResNet34 as the surrogate model and PGD as the candidate adversarial attack. 
The occlusion maps under ResNet50 (f) are based on the transferred adversarial samples from ResNet34. Similarly, for FALcon, we present the final foveated glimpse and the initial fixation point based on the transferred adversarial samples (g). Green solid boxes refer to correct predictions. For the clean samples, FALcon correctly predicts all three instances (d). 
The dark region (1c) aligns with the body of the correct class (tench), resulting in a correct classification for ResNet50. But as indicated in (1f), the dark region shifts and does not align with the body of the object after the injection of adversarial noise leading to an incorrect prediction. For FALcon, although the final foveated glimpse captures the corresponding dark region, the initial fixation point is situated towards the head of the object (1g). This suggests that FALcon was initially primed by more salient features of the object before encountering the probable adversarial patch later. Additionally, downsampling likely reduces the adversity of the patch. For (3f). FALcon similarly to ResNet50 focuses on the dark region (3f) makes an incorrect prediction highlighted by the red box. This suggests that although less vulnerable, there is still room for further improvement for these active vision methods. In the Supplementary, we provide more occlusion maps along with IFPMs.

\section{Conclusion}
\label{sec:conclusion}

In this work we present a robustness study of active vision systems 
under a black-box threat model. We employ various SOTA adversarial attacks and observe consistent and significant performance improvements nearly {2-3} \textbf{x} over passive methods for both GFNet \cite{GFNet_2020} and FALcon  \cite{ibrayev2023exploring}. The enhanced robustness is primarily attributed to two coupled key factors: \textbf{(1)} glimpse-based processing in a downsampled resolution and \textbf{(2)} inference from multiple fixation points. 
Using GFNet we show how (1) mitigates the impact of adversarial noise by downsampling it along with the image. Using FALcon, we show how (2) avoids making mispredictions due to non-uniformity of adversarial noise. We believe this study and its findings will inspire research into incorporating these active methods into standard defense and attack strategies in adversarial learning. This could lead to the development of novel specialized white-box attacks and adaptations of existing defense strategies for active vision methods. In addition to advancing research in this new direction, we anticipate that this study will complement the prevailing literature on bio-inspired defense for adversarial attacks.   

\section{Acknowledgments}
This work was supported in part by, the Center for Brain-inspired Computing
(C-BRIC), the Center for the Co-Design of Cognitive Systems (COCOSYS),
a DARPA-sponsored JUMP center, the Semiconductor Research Corporation
(SRC) and the National Science Foundation (NSF) . We thank Manish Nagaraj and Gobinda Saha (members of Nanoelectronics Laboratory (NRL) at Purdue University) for insightful and helpful discussions.

\bibliographystyle{splncs04}
\bibliography{egbib}

\input{Supplementary}

\end{document}

%% file: Introductions/proposed_introduction.tex
Since the last decade deep learning has seen tremendous progress in a plethora of applications \cite{AlexNet, FasterRCNN, MRCNN, dosovitskiy2021an, GANs, rombach2022high}. However, these networks are susceptible to adversarial inputs, carefully crafted by adding noise that is imperceptible or ignored by human eyes \cite{szegedy2014intriguing, GoodfellowSS14, Papernot2016PracticalBA}. These malicious inputs have been shown to be detrimental to various vision-oriented applications \cite{Athalye2017SynthesizingRA,Hosseini2018SemanticAE,Joshi2019SemanticAA}. To circumvent the effect of such malicious attacks, several defense methods have been proposed, of which Adversarial Training (Adv-T)~\cite{madry2018towards} and Randomized Smoothing (RS)~\cite{RSmoothing} are the most prominent approaches. Although effective against adversarial attacks, the former is computationally expensive while the latter faces scalability issues necessitating the development of other sophisticated methods \cite{andriushchenko2020understanding, CutMix, li2023data, MoosaviDezfooli2018RobustnessVC, yue2023revisiting, zhang2018mixup}. Furthermore, these defense methods are developed purely based on the statistical and computational principles of deep learning. 

Since human eyes are robust to adversarial inputs, there is a research interest in looking at adversarial robustness from the perspective of biological vision. One such direction is to look from the perspective of the human peripheral vision, which processes the visual field in a non-uniform manner. This results in the (central) fovea region, which is processed in high detail, and the (outer) periphery region, which is much less sensitive to fine details and subtle variations. As a result, this line of research is interested in enabling current vision methods with peripheral vision properties, based on the hypothesis that it can allow them to ignore/neglect the effects of adversarial noise. This is explored in the works proposing cortical and retinal fixations \cite{Poggiocortical}, peripheral blurring \cite{shah2023training}, primal visual cortex processing \cite{VOneBlock} or fovea-based texture transformation \cite{gant2021evaluating}. As discussed in R-Blur \cite{shah2023training}, most of these methods model the bio-inspired visual mechanics based on a single fixation point around the input during training.

\begin{figure}[tb]
  \centering
  \includegraphics[height=6.5 cm]{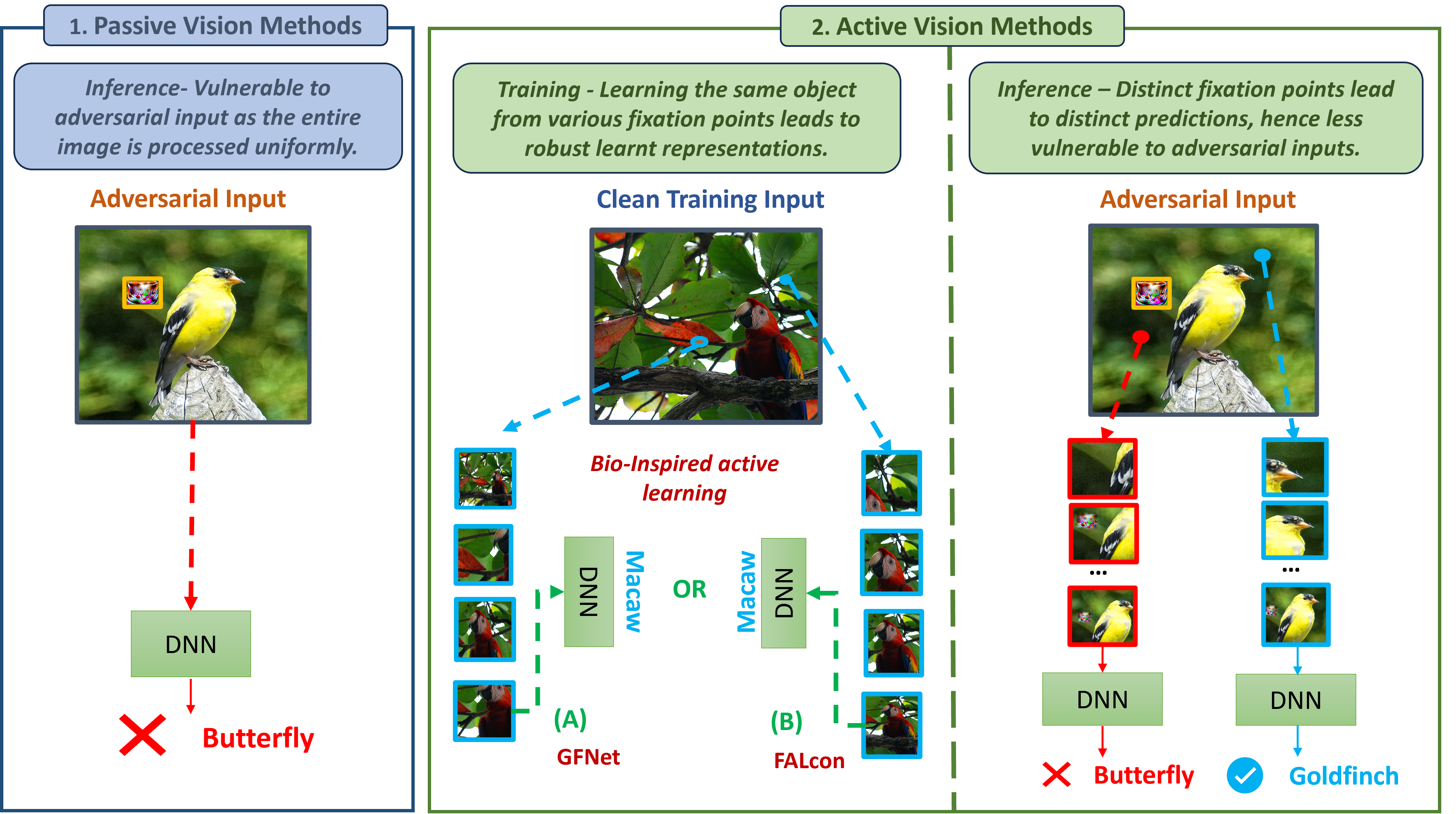}
  \caption{Figure illustrates two methods of processing an adversarial input -- a passive method and an active method. The highlighted yellow box serves as a visual illustration of an adversarial sticker. \textbf{Left column:} One of the probable causes of passive methods being susceptible to adversarial inputs is the uniform processing : that means processing each pixel with same importance. \textbf{Middle column:} In contrast, active vision methods \textbf{(A) GFNet} and \textbf{(B) FALcon} learn the salient features of an object by observing it from multiple distinct fixation points via sequence of glimpses as indicated by the blue boxes. \textbf{Right column:} As a result, during inference, this leads to distinct predictions for the same image not all of which are affected by the adversarial noise. \vspace{-5mm}   
  }
  \label{fig:block}
\end{figure}

On the other hand, it is possible to look at the biological vision from the perspective of an active process~\cite{curcio1990human, land2012animal}. Specifically, human eyes also allow to perceive the visual field from various fixation points (Saccades), causing the change in the non-uniform external resolution across the visual field (Foveation). These mechanisms would actively select where to focus while disregarding redundant or less relevant details in a cluttered environment~\cite{eckstein2011visual}. The idea of enabling deep learning with active vision properties is explored in the works proposing FALcon~\cite{ibrayev2023exploring} and GFNet~\cite{GFNet_2020}. Developed for distinct perception tasks, both demonstrate coupled mechanisms for assessing and focusing on relevant input portions as core components of their learning as shown in \textbf{the middle column of} Figure~\ref{fig:block}. FALcon \textbf{(B)} assigns importance to \textit{predefined} fixation points based on "objectness" and progressively expands until successfully localizing the object or switching to another fixation point. Whereas GFNet \textbf{(A)} determines a relevant fixation point through a glance step and focuses on high-resolution portions through subsequent focus steps, until reaching sufficient confidence in its prediction.

In this work, we advocate that the integration of active vision mechanisms into deep learning can naturally offer robustness benefits. Particularly, we empirically demonstrate that two active vision methods—GFNet and FALcon—exhibit higher levels of adversarial robustness in a black-box scenario compared to passive baseline methods. This is based on the observation that the carefully crafted adversarial noise has a non-uniform distribution across the entire input to match imperceptibility (PGD-like attacks) or size (patch-based attacks) constraints. Hence, as shown in \textbf{the left-most column of} Figure~\ref{fig:block}, the standard passive method of processing inputs, with every pixel having the same importance, can be a probable cause of their adversarial vulnerability. In contrast, owing to the capability to process an input from multiple fixations and through a series of glimpses, active methods are capable of making multiple distinct predictions under the non-uniformity of adversarial noise. For example, as shown in \textbf{the right-most column of} Figure~\ref{fig:block}, FALcon makes two distinct predictions from two fixation points shown by red and blue dots. While the red fixation point results in a misprediction, the prediction from the blue point is consistently robust, illustrating how active vision methods can avoid the adversarial noise\footnote{While an adversarial sticker (generated by P-IFGSM patch-based attack) is used for illustrative purposes, experiments in the paper consider various adversarial attacks.}.

Hence, the main objective of this work is to systematically analyze the robustness of the aforementioned active vision methods. A vanilla black-box threat model is used, where the attacker is unaware of the details of the underlying neural network. The choice is justified based on the goal of evaluating the inherent robustness, where we define it as the robustness when the method is not specifically tailored for adversarial defense. Furthermore, using visualization analysis, the work demonstrates why and how such inherent robustness is achieved using active vision methods. Hence, our contributions are highlighted as follows:
\begin{itemize}
    \item We provide a novel analysis of the inherent robustness of Active Vision Methods in a black box threat model setup.
    \item We empirically demonstrate the superior performance of these methods over standard passive baselines against SOTA adversarial attacks, achieving robustness up to \textbf{$(2-3)$} times as measured by accuracy under attack (\ref{ssec:blackbox}). 
    \item Through both quantitative and qualitative analyses, we showcase the salient learning aspects of these methods that contribute to their inherent robustness. These include glimpse-based learning in a downsampled resolution (\ref{ssec:glimpses}) and inference from distinct fixation points (\ref{ssec:saccades}). 
    \item We show visualizations interpreting how these methods triumph with remarkable resilience against adversarial samples, surpassing their corresponding baselines  (\ref{ssec:saccades})
\end{itemize}


%% file: Tables/InherentRobustness.tex
\def\arraystretch{1.2}%
\begin{table}[t!]
\centering
\caption{Inherent Robustness of Active Vision Methods\vspace{-2mm}}
\label{tab:Main}
\begin{tabular}{c|c|cccc|}
\hhline{~|*5{-}|}
 & \cellcolor{Gray} & \multicolumn{1}{c|}{\cellcolor{Gray}ResNet50 \cite{He2015DeepRL}} & \multicolumn{1}{c|}{\cellcolor{Gray}Adv-T$\star$ \cite{madry2018towards} } & \multicolumn{1}{c|}{\cellcolor{Gray}\textbf{FALcon\cite{ibrayev2023exploring}}} & \cellcolor{Gray}\textbf{GFNet \cite{huang2022glance}} \\ \hhline{~~|*4{-}|}
 
\multirow{-2}{*}{} & \multirow{-2}{*}{\cellcolor{Gray}\textbf{Clean}} & \multicolumn{1}{c|}{\cellcolor{Gray}76.15} & \multicolumn{1}{c|}{\cellcolor{Gray}47.91} & \multicolumn{1}{c|}{\cellcolor{Gray}\textbf{72.97}} & \cellcolor{Gray}\textbf{75.88} \\ \hhline{~|*5{-}|} \noalign{\vspace{0.35ex}} \hline

  
\multicolumn{1}{|c|}{\multirow{2}{*}{Surrogate}} & \multirow{2}{*}{Target} & \multicolumn{4}{c|}{Attack} \\ \cline{3-6}   

\multicolumn{1}{|c|}{} &  & \multicolumn{1}{c|}{PGD \cite{madry2018towards}}   & \multicolumn{1}{c|}{MIM \cite{MIM}}    & \multicolumn{1}{c|}{VMI \cite{VMI}}    & P-IFGSM \cite{GaoZhang2020PatchWise} \\ \hline

\multicolumn{1}{|c|}{\multirow{4}{*}{ResNet34 \cite{He2015DeepRL}}}  & ResNet50  & \multicolumn{1}{c|}{31.46} & \multicolumn{1}{c|}{20.20}  & \multicolumn{1}{c|}{11.61}  & 31.62   \\ \cline{2-6} 

\multicolumn{1}{|c|}{} & Adv-T$\star$ & \multicolumn{1}{c|}{47.63} & \multicolumn{1}{c|}{47.37}  & \multicolumn{1}{c|}{47.16}  & 46.94   \\ \hhline{~|*5{-}|}

\multicolumn{1}{|c|}{} & \cellcolor{Ours}\textbf{FALcon} & \multicolumn{1}{c|}{\cellcolor{Ours} \textbf{49.83}} & \multicolumn{1}{c|}{\cellcolor{Ours} \textbf{37.01}}  & \multicolumn{1}{c|}{\cellcolor{Ours} \textbf{29.40}}  & \cellcolor{Ours} \textbf{40.46}   \\ \hhline{~|*5{-}|}

\multicolumn{1}{|c|}{} & \cellcolor{Ours} \textbf{GFNet}  & \multicolumn{1}{c|}{\cellcolor{Ours} \textbf{57.82}} & \multicolumn{1}{c|}{\cellcolor{Ours} \textbf{48.60}}  & \multicolumn{1}{c|}{\cellcolor{Ours} \textbf{41.17}}  & \cellcolor{Ours} \textbf{46.93} \\ \hline \hline

    
\multicolumn{1}{|c|}{\multirow{4}{*}{ResNet50 \cite{He2015DeepRL}}}  & ResNet50               & \multicolumn{1}{c|}{0.00}  & \multicolumn{1}{c|}{0.00}   & \multicolumn{1}{c|}{0.00}   & 0.00    \\ \hhline{~|*5{-}|}
    
\multicolumn{1}{|c|}{} & Adv-T$\star$  & \multicolumn{1}{c|}{47.68} & \multicolumn{1}{c|}{47.36}  & \multicolumn{1}{c|}{47.22}  & 46.83   \\ \hhline{~|*5{-}|}

\multicolumn{1}{|c|}{} & \cellcolor{Ours} \textbf{FALcon}  & \multicolumn{1}{c|}{\cellcolor{Ours} \textbf{37.54}} & \multicolumn{1}{c|}{\cellcolor{Ours} \textbf{19.96}}  & \multicolumn{1}{c|}{\cellcolor{Ours} \textbf{12.07}}  &   \cellcolor{Ours} \textbf{27.37}      \\ \hhline{~|*5{-}|}
    
\multicolumn{1}{|c|}{} & \cellcolor{Ours} \textbf{GFNet} & \multicolumn{1}{c|}{\cellcolor{Ours} \textbf{51.85}} & \multicolumn{1}{c|}{\cellcolor{Ours} \textbf{42.16}}  & \multicolumn{1}{c|}{\cellcolor{Ours} \textbf{32.33}}  & \cellcolor{Ours} \textbf{43.33}   \\ \hline
    

\end{tabular}
\vspace{-7mm}
\end{table}

%% file: Tables/Horizontal_Geometric_Attack_Token_Grad.tex
\def\arraystretch{1.2}%
\setlength{\tabcolsep}{8pt}
\begin{table}[t!]
\centering
\caption{Geometric Attack using ResNet50 as surrogate.\vspace{-3mm}}
\label{tab:LGV_table}
\begin{tabular}{|cc|c|c|l|c|c|}
\hline
\multicolumn{2}{|c|}{Target} & ResNet50 & Adv-T & CutMix & \cellcolor{Ours}\textbf{FALcon} & \cellcolor{Ours}\textbf{GFNet} \\ \hhline{|*7{-}|}

\rowcolor{Gray}
\multicolumn{2}{|c|}{\textbf{Clean}} & 75.30 & 50.60 & 77.20 & \textbf{75.00} & \textbf{74.20} \\ \hhline{|*7{-}|}

\multicolumn{1}{|c|}{\multirow{2}{*}{\begin{tabular}[c]{@{}c@{}}Attack\\ (LGV +)\end{tabular}}} & PGD & 3.50 & 50.20 & 13.10 & \cellcolor{Ours}\textbf{34.70} & \cellcolor{Ours}\textbf{49.00} \\ \hhline{~|*6{-}|}


\multicolumn{1}{|c|}{} & BIM & 2.40 & 49.80 & 10.00 & \cellcolor{Ours}\textbf{29.60} & \cellcolor{Ours}\textbf{44.70} \\ \hline
\end{tabular}
\vspace{-2mm}
\end{table}
\setlength{\tabcolsep}{6pt} 

%% file: Tables/GlimpseDown.tex
\begin{table}[t!]
\caption{Effect of glimpse-based learning on downsampled resolutions.\vspace{-2mm}}
\label{tab:glimpse_down}
\centering
\begin{tabular}{|c|c|c|cccc|}
\hline
\multirow{3}{*}{Setting} & \multirow{3}{*}{Surrogate} & \multirow{3}{*}{Target} & \multicolumn{2}{c|}{PGD} & \multicolumn{2}{c|}{MIFGSM} \\ \cline{4-7} 
 &  &  & \multicolumn{4}{c|}{Resolution} \\ \cline{4-7} 
 &  &  & \multicolumn{1}{c|}{(96,96)} & \multicolumn{1}{c|}{(128, 128)} & \multicolumn{1}{c|}{(96,96)} & (128,128) \\ \hline \hline

 \rowcolor{Gray}
 &  & ResNet50  & \multicolumn{1}{c|}{52.42} & \multicolumn{1}{c|}{64.42} & \multicolumn{1}{c|}{52.42} & 64.42 \\ 
 \hhline{~~|*5{-}|}

 \rowcolor{Gray}
 \multirow{-2}{*}{1}   & \multirow{-2}{*}{\cellcolor{Gray}\textbf{Clean}} & GFNet  & \multicolumn{1}{c|}{\textbf{75.88}} & \multicolumn{1}{c|}{\textbf{76.70}} & \multicolumn{1}{c|}{\textbf{75.88}} & \textbf{76.70} \\ \hline \hline
    
\multirow{4}{*}{2}     & \multirow{2}{*}{ResNet34}                      & ResNet50              & \multicolumn{1}{c|}{46.03}          & \multicolumn{1}{c|}{54.10}          & \multicolumn{1}{c|}{41.05}          & \textbf{45.73} \\ \hhline{~~|*5{-}|}

    &  & \cellcolor{Ours} GFNet & \multicolumn{1}{c|}{\cellcolor{Ours}\textbf{57.82}} & \multicolumn{1}{c|}{\cellcolor{Ours}\textbf{55.46}} & \multicolumn{1}{c|}{\cellcolor{Ours}\textbf{48.60}} & \cellcolor{Ours}44.72 \\ \cline{2-7} 
    
     & \multicolumn{1}{l|}{\multirow{2}{*}{ResNet50}} & ResNet50 & \multicolumn{1}{c|}{46.00} & \multicolumn{1}{c|}{\textbf{51.41}} & \multicolumn{1}{c|}{41.07} & \textbf{42.31} \\ \hhline{~~|*5{-}|}
     
     & \multicolumn{1}{l|}{} & \cellcolor{Ours} GFNet & \multicolumn{1}{c|}{\cellcolor{Ours} \textbf{51.85}} & \multicolumn{1}{c|}{\cellcolor{Ours} 45.98} & \multicolumn{1}{c|}{\cellcolor{Ours} \textbf{42.16}} & \cellcolor{Ours} 34.76          \\ \hline \hline
     
\multirow{4}{*}{3}  & \multirow{2}{*}{ResNet34} & ResNet50  & \multicolumn{1}{c|}{13.24}  & \multicolumn{1}{c|}{19.64}  & \multicolumn{1}{c|}{8.17} & 12.33          \\ \hhline{~~|*5{-}|}

    &   & \cellcolor{Ours} GFNet  & \multicolumn{1}{c|}{\cellcolor{Ours} \textbf{34.40}} & \multicolumn{1}{c|}{\cellcolor{Ours} \textbf{36.24}} & \multicolumn{1}{c|}{\cellcolor{Ours} \textbf{24.30}} & \cellcolor{Ours} \textbf{24.95} \\ \cline{2-7} 
    
    & \multicolumn{1}{l|}{\multirow{2}{*}{ResNet50}} & ResNet50  & \multicolumn{1}{c|}{0.30} & \multicolumn{1}{c|}{0.13} & \multicolumn{1}{c|}{0.35} & 0.16 \\ \hhline{~~|*5{-}|}
    
    & \multicolumn{1}{l|}{}  & \cellcolor{Ours} GFNet & \multicolumn{1}{c|}{\cellcolor{Ours} \textbf{17.96}} & \multicolumn{1}{c|}{\cellcolor{Ours} \textbf{12.12}} & \multicolumn{1}{c|}{\cellcolor{Ours} \textbf{10.63}} & \cellcolor{Ours} \textbf{6.80} \\ \hline
\end{tabular}
\vspace{-6mm}
\end{table}

%% file: supplementary.tex
\appendix

\section{Active Vision Methods}
\label{sec:avm_app}

\subsection{Background}
In this section, we offer a high-level overview and insights into the learning and inference of two candidate active vision methods — Glance and Focus Networks (GFNet) \cite{GFNet} and FALcon \cite{ibrayev2023exploring}. Both of these methods incorporate human-inspired mechanisms like foveation and saccades. They simulate foveation by cropping glimpses from the image based on fixation points, rather than blurring the extracted glimpses as observed in human peripheral vision. Consequently, this cropping can be viewed as foveation with extreme cut-off.   

\subsection{Glance and Focus Networks}
\label{ssec:glfnet_app}

\subsubsection{Overview}

The primary objective of this work was to introduce a vision system inspired by the human perceptual system capable of disregarding redundant spatial information, instead focusing solely on the object-relevant details. The learning task is that of image classification within a constrained computational budget and is modeled as a sequential decision process. The broad overview of the two stage framework, glance and focus is presented in Figure \ref{fig:gfblock_app} and the corresponding highlights in the subsequent sections.

\subsubsection{Learning}

\begin{figure}[tb]
  \centering
  \includegraphics[height=7.0 cm]{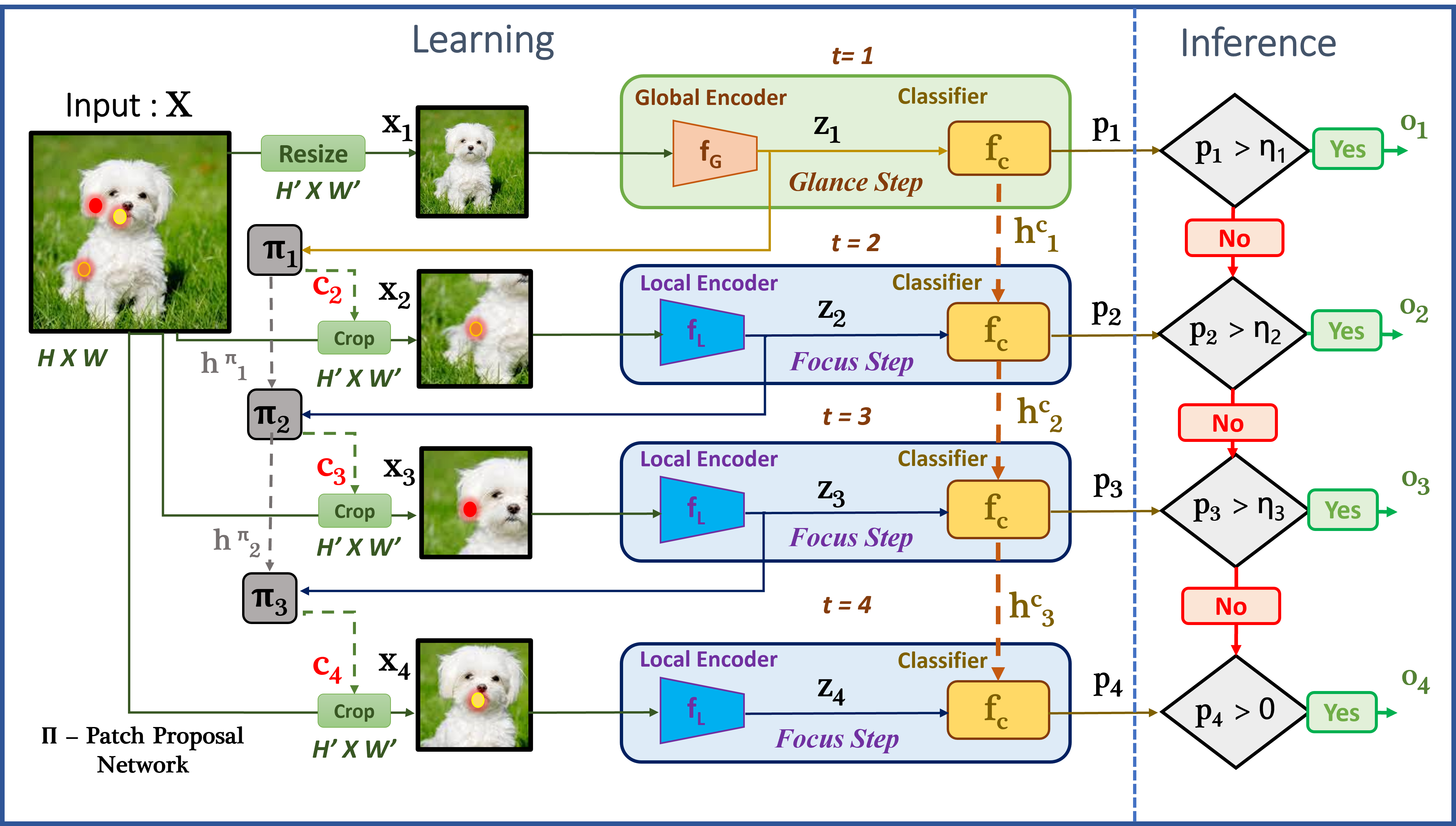}
  \caption{(\textbf{Learning \& Inference}) The figure provides an overview of GFNet's operation. It begins by downsampling the input image to a lower resolution for rapid prediction $(p_{1})$, termed \textbf{$f_{G}$ (Glance)} at $t=1$. If the network lacks confidence $(p_{1} < \eta_{1})$, it enters subsequent \textbf{$f_{L}$ (Focus)} steps until certainty is attained or till $(t=4)$. Each focus step analyzes a patch $(H' \times W')$ cropped from the original input $(H \times W)$ centered around $(c_{t})$ illustrated by colored dots. These co-ordinates are determined by a patch proposal network \textbf{$\pi$}. The process is depicted for a sequence length of $4$.}
  \label{fig:gfblock_app}
\end{figure}

The learning occurs in two distinct steps -- a glance step and subsequent multiple focus steps. In the diagram, we illustrate the learning till $T = 4$ which is the maximum length of the sequence. Each subsequent step is denoted by $t$. The task at hand is to correctly classify an input image $X$ within a given number of time steps.

In the first step, the image with full resolution $(H \times W)$ is downsampled to a much lower resolution $(H' \times W')$ where $H' < H$ and $W' < W$, and then processed to make a quick prediction based on the globally processed information through a global encoder $f_{G}$. This is known as the \textit{Glance} Step, akin to how humans tend to make quick predictions by glancing through images. Along with the prediction $p_{t}$, a patch proposal network $\pi_{t}$ predicts a region proposal $(\pi_{t})$ based on the most class-distinctive region for the subsequent step $(t+1)$ to process. The output of this network consists of the location of each image patch, which comprises normalized image coordinates corresponding to the center coordinates of each patch $(c_{t+1})$. These image co-ordinates are the fixation points for the subsequent glimpses. 

At $t=2$, as illustrated in Figure \ref{fig:gfblock_app} there is an orange fixation point $(c_{2})$ around the leg region of the dog which is the predicted co-ordinate at that time step. Using these coordinates, the corresponding patch is cropped from the full image at size $H' \times W'$ and fed to the local encoder $f_{L}$. Since the cropped image is a small patch entailing sharp details, this step is known as \textit{Focus} step. This step is repeated until the maximum length of the input sequence, which is $T=4$ in this case. During each iteration, the local encoder is progressively trained on the smaller patches based on the fixation points determined by the patch proposal network. The sequential glimpses are illustrated in the diagram, centered around distinct fixation points per step as given by $c_{2}, c_{3}$, and $c_{4}$ for time steps $2, 3$, and $4$ respectively. Thus, at each focus step, the network observes different class-discriminative regions of the object with sharper details to successfully classify the image. 

The global and local encoders $f_{G}$ and $f_{L}$ respectively are deep convolutional neural networks and the classifier $f_{C}$ is a deep recurrent neural network. The classifier $f_{C}$ aggregates the information from all previous inputs $({h^{c}}_{t-1})$ and the subsequent feature maps $z_{t}$ from the encoder to produce a prediction at each step. $f_{G}$ , $f_{L}$ and $f_{C}$ are trained simultaneously in a supervised manner to produce correct predictions $p_{t}$ at high confidences for the entire length of input sequence from $t=1$ to $t=4$ as depicted in the diagram. 

The patch proposal network is also a recurrent neural network trained via a policy gradient algorithm to select the patches that maximize the increments of the softmax prediction on the ground truth labels between adjacent two steps. The inputs to this patch proposal network are the previous hidden representations and subsequent feature maps, denoted as $h^{\pi}_{t-1}$ and $z_{t}$ respectively. 

Thus, the network learns the representations of each object by actively predicting class-discriminative image patches centered around the fixation points and learning the same, which consist of important and salient details.  

\subsubsection{Inference}
GFNet performs inference in two distinct steps -- a glance step and subsequent multiple focus steps (Figure~\ref{fig:gfblock_app}). In the glance step, the full-resolution image $(H \times W)$ is first downsampled to a much lower resolution $(H' \times W')$. Then it is passed through a global encoder $f_{G}$ to make a swift prediction based on the global features.  If the confidence $p_{t}$ exceeds the threshold $\eta_{t}$, where $\eta$ is a pre-defined threshold 
\cite{huang2018multiscale, yang2020resolution}, the process halts. A patch-proposal network evaluates these features concurrently to predict the fixation point for the subsequent focus step. Each patch is generated based on the most salient features of the object centered around the fixation point. This is denoted by the orange dot in the second patch. This $H' \times W'$ patch is cropped from the image and inputted into the local encoder $f_{L}$. Simultaneously coordinates for the next focus step are produced by the patch-proposal network. The iterative process persists until the network gains sufficient confidence in its prediction, or reaches the end of the sequential process $t = T$. To understand the robustness aspect through inference on transferred adversarial samples, we keep the early termination inactive, allowing inference to continue until $t = T$. This provides the network distinct fixation point, at each step to assess the input in the presence of adversarial noise. The global and local encoders $f_{G}$ and $f_{L}$ respectively are deep convolutional neural networks and are trained on the low dimensional inputs of $H' \times W'$.    

\subsection{FALcon}

\begin{figure}[tb]
  \centering
  \includegraphics[height=6.50 cm]{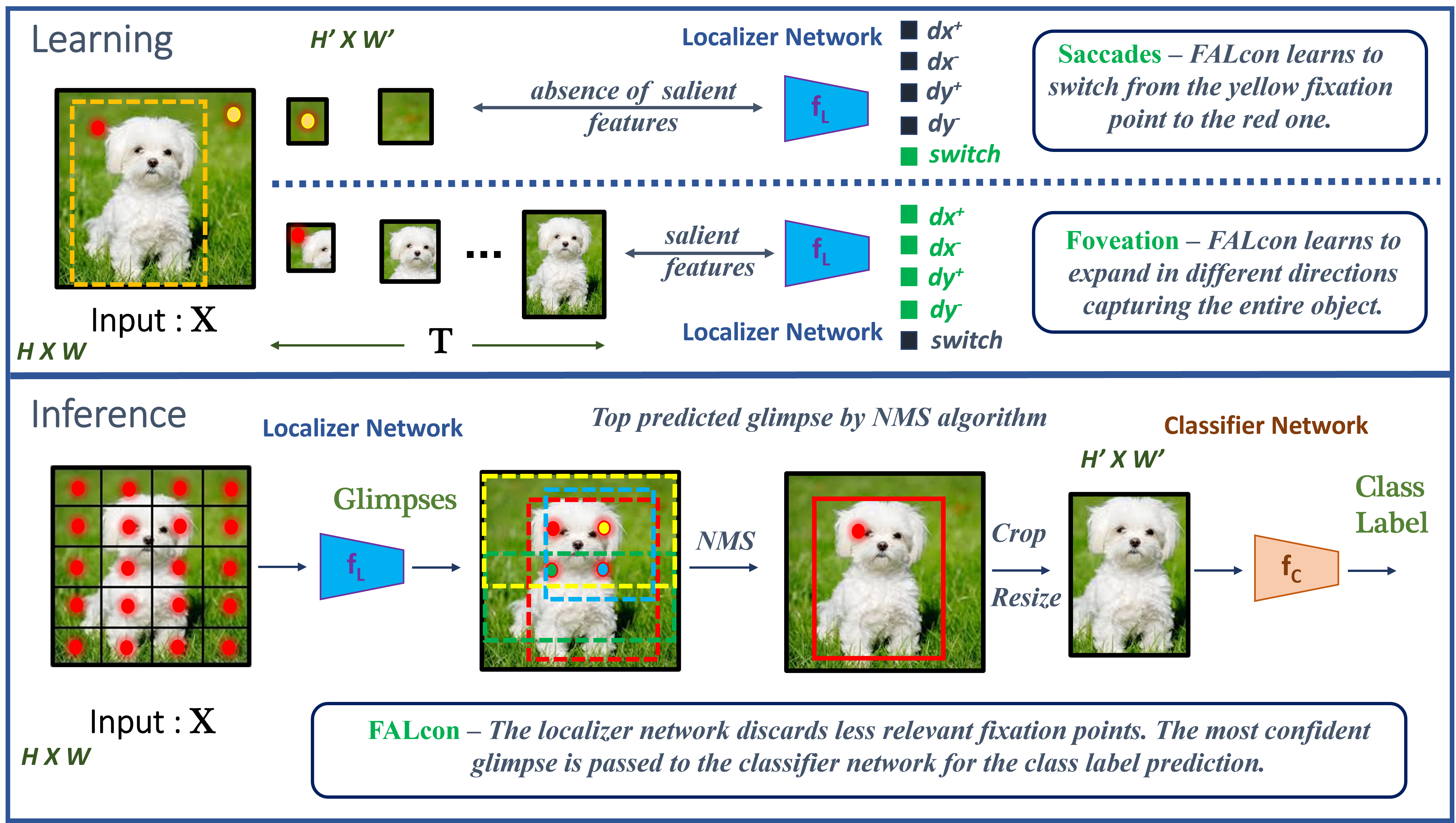}
  \caption{The figure provides a high-level overview of FALcon. During \textbf{Learning}, a Localizer network $(f_{L})$ is trained to predict five distinct actions (four for expansion and one for switching), enabling it to learn the importance of each fixation point illustrated by colored dots. Learning occurs in a downsampled resolution of $(H' \times W')$. During \textbf{Inference}, $f_{L}$ starts from each pre-defined multiple fixation point (20 red dots). If salient object features are present, $f_{L}$ performs the learned expansions to capture the object (4 colored dashed boxes, colored dots). The most confident final foveated glimpse (red solid box) is cropped $(H' \times W')$ and presented to the classifier $(f_{C})$. }
  \label{fig:falcon_app}
\end{figure}

\subsubsection{Overview}
The main objective of this work \cite{ibrayev2023exploring} was to devise an active vision method equipped with human-inspired mechanisms such as foveation and saccades and then successfully apply this method to the task of Weakly Supervised Object Localization. The aim is to produce bounding box predictions by learning solely from image level information such as class labels with pseudo box \footnote{Pseudo bounding boxes are annotations produced by any oracle method and are not ground truth box annotations.} annotations during training. Although the method outputs bounding box annotations as final predicted glimpses along with labels, for our evaluation purposes, we only utilize the label prediction.

\subsubsection{Learning}
In the upper part of Figure \ref{fig:falcon_app}, we present a high-level overview of the learning stage of FALcon. The learning algorithm directs a localizer network $f_{L}$ to produce a sequence of foveated glimpses using two learned techniques: foveation and saccades. An initial fixation point marks the starting position of the expansion process. A foveated glimpse is defined as a cropped region from the image (initiated at fixation points) with dimensions dictated by the foveation expansion process, as described later.  Each foveated region is downsampled to lower dimensions of $H' \times W'$. For localization, given an input image $X$, an initial fixation point and a pseudo-bounding box (illustrated as an orange dashed box), the localizer $f_{L}$ learns five distinct actions, as depicted in the diagram, to produce a sequence of foveated glimpses on which the network is trained. This is a sequential iterative process carried out for $T$ steps. The glimpse at the last step of the iterative process is defined as the final foveated glimpse. The dimensions of the final foveated glimpse are designated as the dimensions of the predicted bounding box.

The localizer begins its process with an initial fixation point (depicted as a red point) within the orange pseudo bounding box. Starting with a predefined resolution for the initial glimpse, the localizer then iterates through four potential expansions: rightward $(dx^{+})$, leftward $(dx^{-})$, downward $(dy^{+})$, and upward $(dy^{-})$, with the top-left corner of the initial glimpse as the origin $(0,0)$. These expansions aim to emulate the foveation process, wherein the model actively adjusts its focus to capture salient object features. The model's decision to expand in each direction is based on its confidence in capturing an object, with expansions guided by a fixed glimpse step size. Importantly, the expansions are constrained by the pseudo bounding box and the relative position of the initial fixation point to prevent glimpses from extending beyond this boundary. At each iteration $t$, the target for expansion is determined based on the dimensions of the current glimpse and the reference pseudo-bounding box. For instance, if the target is (${{1,0,1,0}}$), the model prioritizes expanding to the right and downward given the bounding box's location and the current glimpse. This iterative process continues for a fixed length of the foveation sequence, denoted as $T$, until the localizer successfully captures the entire object.

During training, FALcon undergoes a learning process where it evaluates the relevance of a fixation point and decides whether it needs to switch to another fixation point. If the initial fixation point falls outside the pseudo bounding box (orange point), indicating a potential need for a switch to focus on relevant features, the network is guided to make the switch (switch target is \textit{true}). Conversely, if the network determines that a switch is necessary based on the absence of salient features in the current glimpse sequence, a new fixation point within the pseudo box is presented, and no switch is required (switch target is \textit{False}). This learning mechanism simulates the way human vision actively adjusts focus, akin to saccadic movements when essential features are not detected in the current field of view.

By learning the relevance of different fixation points, FALcon learns to capture objects using various trajectories of foveated regions. This ensures the diversity of learned representations of salient object features in an image. The localizer is a deep convolutional neural network, and the five actions are trained based on binary cross-entropy loss between predictions and targets per iteration ${t}$ in the input sequence of length ${T}$.  

\subsubsection{Inference - Top Prediction}
During inference, the input image $X$ is divided into grid cells, as illustrated in the first image at the bottom part of Figure~\ref{fig:falcon_app}. Each grid cell is considered as an initial fixation point (red dot). The localizer $f_{L}$ initiates from each point with a pre-defined glimpse size to inspect salient object features. In the absence of salient features, the localizer deems there is no potential object and hence switches to another fixation point. If the localizer encounters any relevant features and hence assumes the presence of a potential object, it expands from the initial glimpse dimensions in four independent directions to generate a sequence of foveated glimpses. Based on the expansion operations, each foveated glimpse is cropped from the input image at a downsampled resolution of $H' \times W'$. The sequence of expanding for a potential fixation point continues until the final foveated glimpse captures the entire object. This process iterates for all potential points that could lead to various final foveated glimpses for the same object. This is illustrated by the colored dots and dashed colored boxes in the second image. For object localization, these final foveated glimpses undergo non-maximum suppression to yield the most confident prediction indicated by the solid red bounding box in the third image. The region corresponding to the most confident final foveated glimpse is cropped at dimensions $H' \times W'$, and then presented to the classification network $f_{C}$ for class prediction. In our evaluation, we only use this final class prediction label. The remaining potential dots, except the most confident one although not utilized for localization, serve as a map to understand the effect of non-uniform adversarial noise injected into images. In our experiments, we visually demonstrate how the noise affects some of these fixation points, but not all resulting in enhanced robustness. Both localizer $f_{L}$ and classifier $f_{C}$ are deep convolutional neural networks. During learning, $f_{C}$ is not fine-tuned on the lower dimensional inputs as is done for GFNets. 

\subsubsection{Inference - Any Predictions}
\label{sssec:descany}
In the previous section, we discussed the use of multiple fixation points for inference and highlighted our choice of selecting the most confident prediction, determined by the NMS algorithm, as our final output. This approach ensures a fair comparison for image classification since passive baseline methods typically provide only one prediction (Top-1). Alternatively, we could consider all potential predictions sorted by NMS based on class confidence and evaluate the ordered list. Here, any prediction matching the ground truth is deemed correct.  We show quantitative results in Section (\ref{sssec:AnyPreds}) by augmenting Table 1 already present in the main paper. FALcon exhibits an improvement of nearly $4\%$ over the Top case when the constraint is removed unlocking its true potential. The rationale behind this improvement is as follows: The NMS algorithm outputs an ordered list sorted by class confidence. Referring to Figure \ref{fig:falcon_app}, the remaining potential dots (green, yellow, blue) also have the potential to yield correct class predictions when fed into the classifier $f_{C}$. While this might not be particularly advantageous for clean samples, the scenario changes significantly for corresponding adversarial images. We have visually demonstrated the effectiveness of predicting from multiple fixation points using Initial Fixation Points Maps (IFPM). These maps reveal how non-uniform adversarial noise impacts various initial fixation points differently. Adding adversarial noise alters the features in an image leading to incorrect predictions. Consequently, some fixation points may lead to more confident incorrect predictions, while others, although less confident, might still be correct. This advantage of inferring from multiple fixation points isn't fully captured when assessing the most confident prediction based on the NMS sorted list. Considering \textit{any prediction} from the NMS list as correct addresses this gap. Presenting these additional results underscores the benefits of inferring from multiple fixation points. However, it's important to note that in this setup, the assumption persists that there is only one correct object per class, as dictated by the image classification task.

\subsubsection{Inference - Multiple Objects}
\label{sssec:descmulti}
For a dataset such as ImageNet \cite{ImageNet}, there are multiple instances where there is more than one object per image. Another added benefit of FALcon, as presented in the original paper, \cite{ibrayev2023exploring} is that it can detect multiple objects. The reason behind this is, once again, the inference from distinct fixation points. This is beneficial for a black-box scenario, where the underlying architecture although trained for image classification has the inherent property of detecting more than one object per image. This is illustrated pictorially in Section (\ref{sssec:Multi}), via Initial Fixation Point Maps (IFPM) for multiple objects. In scenarios with multiple objects, there can be more than one correct class per image. This extends the Any Predictions Case, where during the evaluation of the ordered list, any prediction matching the ground truth of any object in an image is considered correct. However, it's crucial to acknowledge that even in this setup, the assumption remains that there is at most one correct prediction per image. This prediction can correspond to any one or to all of the classes present in that image.

\section{Additional Experimental Results}

\subsection{Inherent Robustness in the Black-box Setup}

We generate adversarial samples from surrogate classifiers \(S_{i}\) using iterative adversarial attacks such as PGD \cite{madry2018towards}, MIFGSM \cite{MIM}, VMIFGSM \cite{VMI} and Patchwise-IFGSM \cite{GaoZhang2020PatchWise}. The general iterative attack setup can be defined as :

\begin{equation*}
    x^{adv}_{t+1} = x^{adv}_{t} + \alpha*sgn(\nabla_x J((x),y_{\text{true}})) \mid S_{i} (x^{adv}_{t}) \neq y_{\text{true}}
\end{equation*}

Based on each attack, the expression \( \nabla_x J((x),y_{\text{true}}) \) can be modified according to the method employed by each attack. The attack hyper-parameters include the number of iterative steps \( t \), \( \alpha \) representing the step size or perturbation magnitude applied at each iteration to update the input \( x \) towards maximizing the loss, and \( \epsilon \), which is the maximum allowable budget for the attack.

We provide additional results in this section with \textbf{DenseNet121} \cite{Densenet} as the surrogate model. \acc is provided as the metrics for comparison. 

\input{Tables_App/Main_Table_DenseNet121}

\textbf{Results:} Presented in Table \ref{tab:Main_densenet}, we observe consistent trends across non-ResNet surrogates ($S_{i}$). Active vision methods demonstrate superior accuracy (\acc) compared to passive baseline methods, ResNet34 and ResNet50. Additionally, we include a ResNet50 trained via CutMix \cite{CutMix}, which exhibits inferior robustness compared to the active vision methods, except for P-IFGSM \cite{GaoZhang2020PatchWise}, where similar robustness is observed with FALcon. It's noteworthy that both active vision methods can be further enhanced with the regularization technique from the original CutMix paper, contributing to their robustness.


\subsection{Token Gradient Regularization} 
\label{sssec:tgr}

TGR \cite{zhang2023transferable} introduces a gradient based transfer attack algorithm for Vision Transformers (ViT) \cite{dosovitskiy2021an} and its variants such as Class-Attention in Image Transformers (CaiT) \cite{Touvron2021GoingDW} and Pooling based Vision Transformer (PiT) \cite{Heo2021RethinkingSD}. The method removes tokens with extreme values and reduces variance of back-propagated gradients. It utilizes token gradient information from both the Attention and Query-Key-Value components within an attention block, as well as from the MLP component within the MLP block, to generate adversarial samples.

\begin{align*}
    Grad_{adv} = TGR(Grad_{QKV}, Grad_{Att}, Grad_{MLP}, k, s) \\
    x^{adv}_{t+1} = x^{adv}_{t} + \alpha*sgn({Grad_{adv}})
\end{align*}

Here k denotes the top-$k$ or bottom-$k$ input gradients with highest and lowest values respectively which denote the extreme tokens. $s$ is a scaling factor for the gradients and $\alpha$ is a hyper-parameter to control the step size. This method is effective against CNN models as well.

\textbf{Setup} We follow the experimental setup as outlined in the original paper \cite{zhang2023transferable} and show results on a test set of random $1000$ images from the ImageNet\cite{ImageNet} validation set. Table (\ref{tab:TGR-table}) displays the {\acc}  of different target networks on adversarial samples transferred from different surrogate architectures. Alongside the baseline methods previously studied, we investigate the inherent robustness of several notable vision transformer architectures in image classification such as Swin V2 Transformer, \cite{SwinV2}, Focal Modulation Net \cite{yang2022focal} and Robust Vision Transformer \cite{Mao2021TowardsRV}. The vision transformers examined are their corresponding tiny versions, with a parameter count close to 25 million parameters, comparable to the CNN baselines studied. We adopt the implementation of Vision Transformers (ViT-B/16) \cite{dosovitskiy2021an} and their variants PiT-B, and CaiT-S/24 \cite{Touvron2021GoingDW, Heo2021RethinkingSD} as \(S_{i}\) with pre-trained weights from PyTorch.

\textbf{Results} The top performing CNN and Transformer models for each \(S_{i}\) are indicated in bold. The oracle methods, Adv-T$\star$ and Adversarially trained Robust Vision Transformer Adv-RVT$\star$ provide the best case for CNNs and Transformers respectively. Among transformers, Robust Vision Transformer (RVT) \cite{Mao2021TowardsRV} and Focal Net \cite{yang2022focal} exhibit higher \acc in two instances among the vision transformers. This can be attributed to the robust feature representation learnt for RVT and the human inspired feature-based attention learning approach for Focal Net which helps in foreground-background segregation respectively. 

\input{Tables_App/TGR_multiple_pred}

Notably, GFNets exhibit superior performance among CNNs, with an average accuracy increase of approximately 5.0\% across the three distinct surrogates compared to a passive ResNet50. For CaiT-based samples, both FALcon and GFNet demonstrate superior performance across all non-adversarially trained models (CNNs and Transformers). Despite FALcon showing slightly lower standard accuracy on the clean 1000 random test set, its relative drop in accuracy on samples generated by $S_{i}$, ViT-B/16 is comparable to ResNet50's. Moreover, this relative drop is even lesser for CaiT-S/24 and PiT-B respectively, suggesting that the inherent robustness properties of active vision methods transfer to transformer-based adversarial attacks as well. Furthermore, fine-tuning FALcon's \(f_{C}\) on downsampled images might help bridge the performance gap with GFNet to some extent.



\subsection{Effect of distinct fixation points (case study: FALcon)}

\input{Tables_App/Any_Preds_big_table}

In this section, we delve into additional results on FALcon's inference for Any-Predictions (\ref{sssec:descany}) and Multiple Objects (\ref{sssec:descmulti}). While the efficacy of performing inference from multiple fixation points is detailed in the main manuscript, these quantitative results further support this claim. To maintain clarity and consistency, we excluded any and multiple predictions from the main manuscript, as it focuses on a fair comparison with passive baseline methods, which typically produce one prediction per sample.


\subsubsection{Any Predictions}
\label{sssec:AnyPreds}
The explanation of FALcon making any predictions is detailed in Section (\ref{sssec:descany}). Here, we present the results in Table (\ref{tab:Main_Any_Preds}) based on Any Predictions by simply adding FALcon-Any to Table 1 from the main manuscript. 

\textbf{Results} As shown in Table (\ref{tab:Main_Any_Preds}), FALcon-Any demonstrates an improvement of nearly $4\%$ for both surrogates over the Top case when the constraint of utilizing the most confident prediction based on NMS is removed. This highlights how evaluating based on any predictions unlocks FALcon's true potential.

\subsubsection{Multiple Objects} 
\label{sssec:Multi}
\begin{figure}[t!]
  \centering
  \includegraphics[height=6.0 cm]{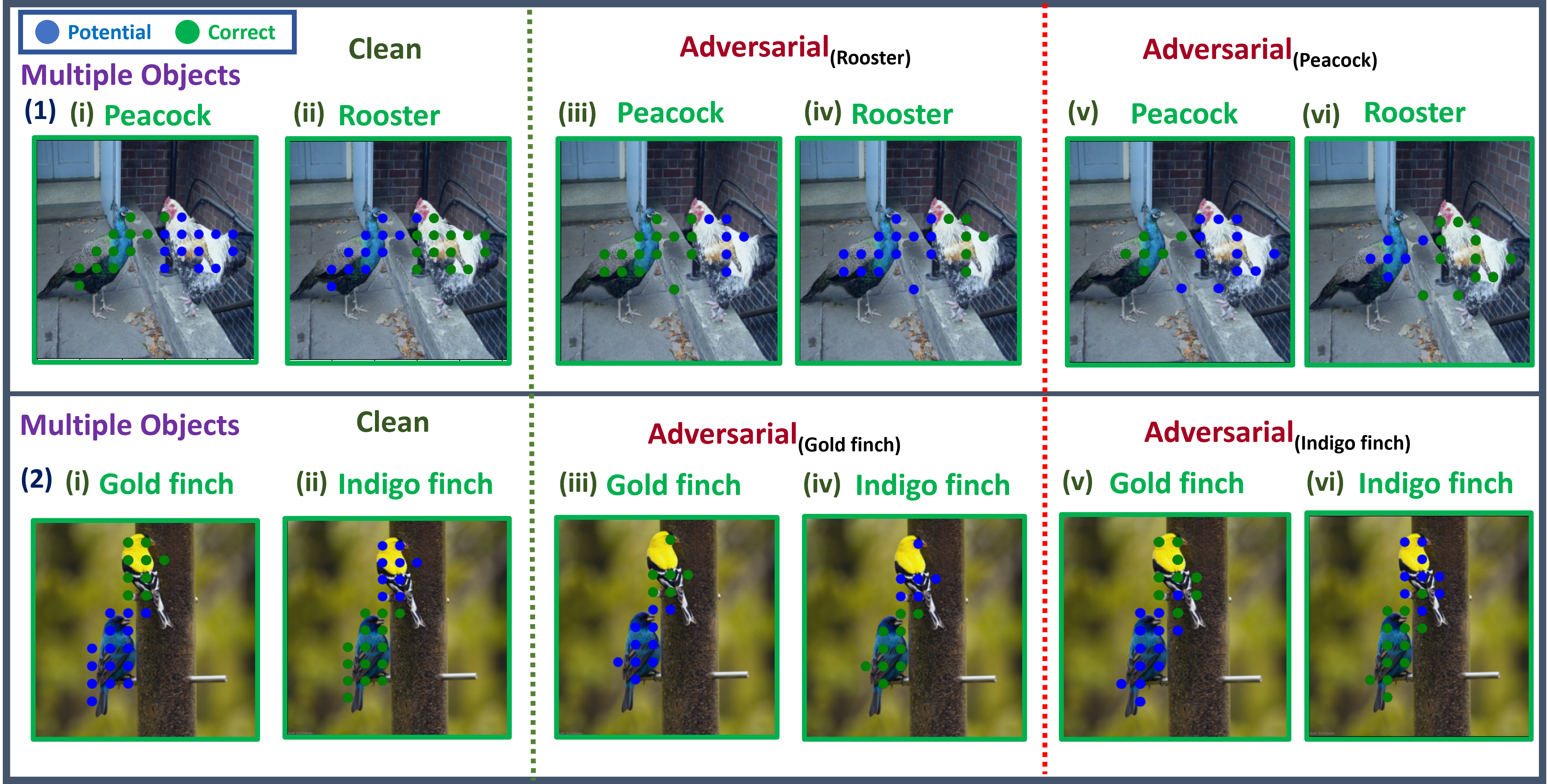}
  \caption{Figure illustrates Initial Fixation Point Maps \textbf{(IFPM)} for multiple objects. For \textbf{(1)}, the reduction in  \textcolor{OliveGreen}{correct points} is illustrated as \textbf{(i to v)} for the class Peacock. Similarly, for class Rooster, this is illustrated as \textbf{(ii to iv)}. \textbf{(iii \& vi)} show the effect on \textcolor{OliveGreen}{correct points} for a class when attack is generated based on the other class.  
  }
  
  \label{fig:FPM_image_Multi}
\end{figure}

FALcon \cite{ibrayev2023exploring} offers the added advantage of detecting multiple objects, facilitated by the inference from distinct fixation points. In a black-box scenario, where the underlying architecture is unknown, a model trained for image classification but can inherently detect multiple objects per image, can prove to be beneficial. The setup and explanation are detailed in Section (\ref{sssec:descmulti}). Illustrated in Figure (\ref{fig:FPM_image_Multi}), Initial Fixation Point Maps for multiple objects depict this concept, building upon the explanation provided for single objects in the main manuscript. 

In \textbf{(Sample 1)}, featuring two classes, peacock and rooster, we individually generate non-targeted adversarial samples for each class, such as $Adversarial_{peacock}$ for the peacock class. The non-uniform distribution of adversarial noise affects fixation points differently, leading to a reduction in correct points from the clean sample to the adversarial one \textbf{(For Peacock (i to v)} and \textbf{(For Rooster (ii to iv))}. However, despite the introduction of adversarial noise, the number of correct fixation points for the correct class remains consistent (iv \& v).

\textbf{Sample (2)} In the second sample with goldfinch and indigo finch, the decrease in correct points for goldfinch is evident from \textbf{(i to iii)}, while for indigo finch, it is observed from \textbf{(ii to vi)}. Despite a slight variation in order between the two samples (pictorially), the generation process remains consistent.

Table \textbf{\ref{tab:multiple_pred}} quantifies this observation. FALcon-Multi exhibits a slight increase of $0.5\%$ compared to FALcon-Any (shown in Table \ref{tab:Main_Any_Preds}) for PGD. However, for a stronger attack like MIFGSM, there is hardly any improvement. As mentioned in Section \ref{sssec:descmulti}, it's important to note that even in this scenario, the assumption remains that there is at most one correct prediction per image, which can correspond to any one or to all of the classes present in that image. Hence the relative improvement from the Any case.

\subsubsection{Occlusion Maps: Initial Fixation Point Variations} 
\label{sssec:Occ_Var}

\begin{figure}[t!]
  \centering
  \includegraphics[height=5.5 cm]{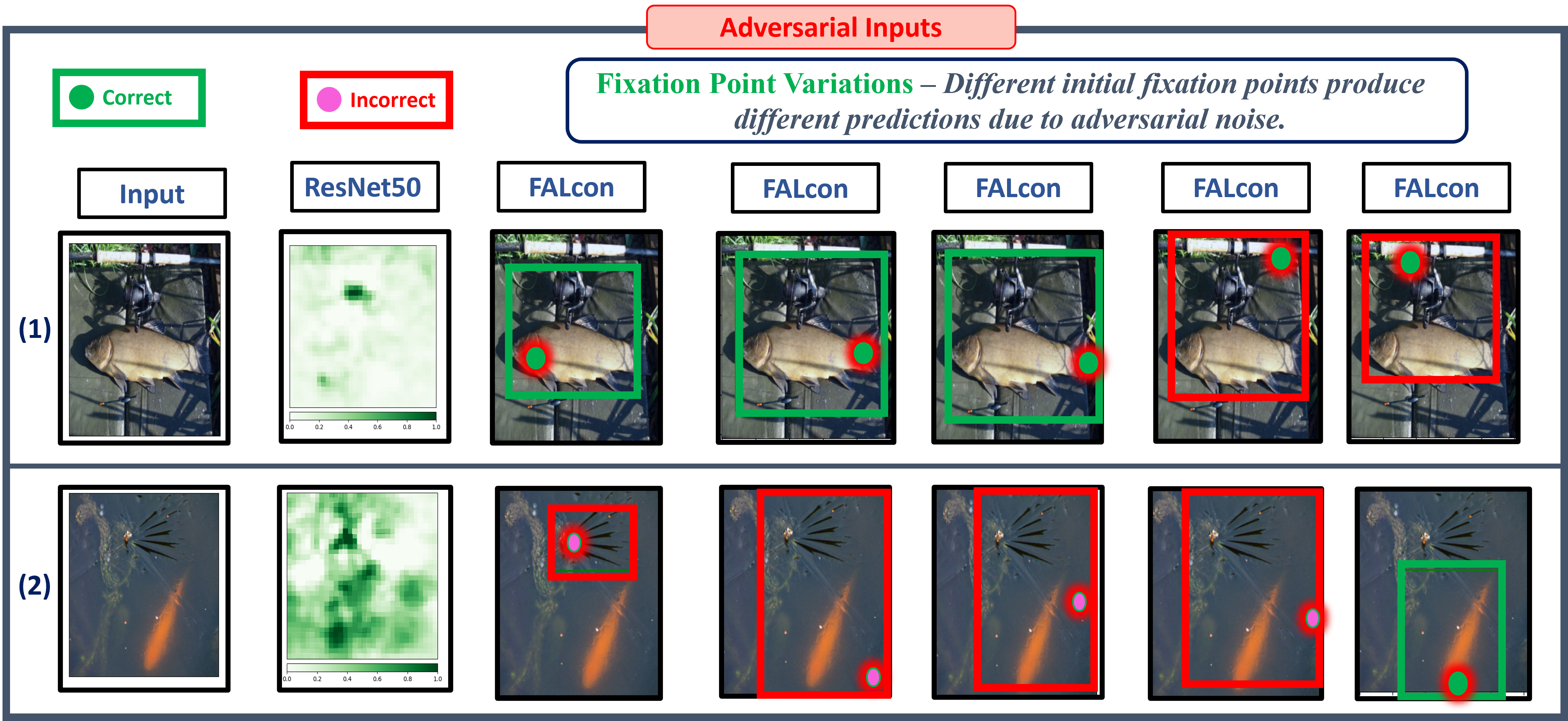}
  \caption{The figure illustrates how the final foveated glimpse varies with different fixation points in the presence of adversarial noise. In sample (1), most predictions are correct (solid green box), while in sample (2), the opposite is observed (solid red box). This variation is due to the relative position of the initial fixation point and the prominent regions influenced by adversarial noise (as viewed from the occlusion maps for ResNet50), affecting model predictions. This demonstrates how active vision methods are primed based on their initial fixation point. 
  }  
  \label{fig:occl_fix}
\end{figure}

In this section, we illustrate how changing the initial fixation points can influence FALcon to produce diverse final foveated glimpses. This concept relates to Figure 1 in the main manuscript, showcasing how observing the same object from various fixation points can influence the model, depending on the point's relative position and the uneven distribution of adversarial noise.  The visualizations in Figure \ref{fig:occl_fix} provide additional insight into the model's behavior in this context.

\textbf{Setup} The second column displays occlusion maps generated with ResNet50 on transferred adversarial images from ResNet34, as previously demonstrated in the main manuscript. The first image under FALcon displays the original final foveated glimpse following FALcon's inference pipeline. For the subsequent images, to understand model behavior on adversarial images, we fix the initial points as depicted and perform inference solely based on these points. The final foveated glimpse is presented inside the solid bounding boxes.

In Sample (1), positioning the fixation points along the object's body helps FALcon focus on the important features first, then notice the adversarial dark spot. This primes FALcon in making the right prediction. However, initialing the fixation point on either side of the dark spot, as indicated by the occlusion map, leads the model to make wrong predictions, as shown by the red solid box.

In Sample (2), the adversarial noise is spread unevenly across the entire image, unlike the previous sample. Fixing the points at different parts of the image leads to incorrect predictions, as indicated by the red solid boxes. Only in the last image, FALcon predicts correctly based on the fixation point near the head of the goldfish. This experiment highlights the potential for rectifying predictions by suggesting safe fixation points. Further exploration in this direction could yield interesting insights.    

The fixation points and the bounding boxes are scaled up to enhance clarity. 

\section{Additional Related Work}

\subsection{Two Stream and Applications}

The "two-stream hypothesis" from neuroscience models the visual processing in the human visual cortex as two separate sections (ventral and dorsal) in the brain catered towards answering two key questions related to visual stimuli: \textbf{what} (ventral) and \textbf{where} (dorsal). Recently, FABLE \cite{ibrayev2024twostream} proposed a localization framework modeling the ventral stream as a supervised feature extractor model and the dorsal stream as a separate model trained via reinforcement learning. The network was additionally shown to generalize to samples of learnt classes from different datasets that were not used during training. Chun \etal \cite{tao2024image} built on top of this framework with a bi-directional LSTM to understand and learn proper sequential semantics in objects. This method makes the implicit order and structure in images explicit, by learning basic parts of an object forming the entirety of the object.  

\subsection{Transformers and Robustness} 

Swin V2 \cite{SwinV2} is an updated version of Swin Transformer \cite{SwinV1} in terms of model capacity and the ability to process inputs at higher resolutions. The paper \cite{FocalTransformer} introduces focal attention, a novel mechanism that combines fine-grained local and coarse-grained global interactions to efficiently capture short- and long-range visual dependencies in Vision Transformers. Leveraging focal attention, the proposed Focal Transformers achieve superior performance compared to state-of-the-art Vision Transformer models across multiple image classification and object detection benchmarks. Focal Modulation Networks \cite{yang2022focal} introduced a focal modulation mechanism that flips the conventional self-attention mechanism, aligning more closely with human-like feature-based attention rather than spatial attention. This approach, demonstrated improves over Focal Transformers \cite{FocalTransformer}, and enhances object segmentation by focusing on features rather than spatial positions. 

The \textbf{Robust Vision Transformer} \cite{Mao2021TowardsRV} identified weaknesses in transformer models for adversarial robustness and introduced novel techniques like position-aware attention scaling and patch-wise augmentation to enhance robustness across various shifts, resulting in a more resilient vision transformer.  Qin et al. cite{qin2021understanding} provided heavy experimentation to show how negative patch-based augmentation can improve robustness of vision transformers. Shao et al. \cite{shao2022on} provided insights such as ViTs are less inclined towards high-frequency components in an image that contain spurious correlations. This makes them more robust against high-frequency perturbations. Wei et al. \cite{wei2022towards} proposed two techniques that leverages both pacthes and self-attention to improve the transferability of adversarial attacks on vision transformers. Token Gradient Regularization \cite{zhang2023transferable} is a SOTA transformer based 
transfer attack that removes extreme tokens from the backpropagation update to mitigate inter block variance in transformers. This makes the attack more general and eliminates vulnerability to transformer-specific local attacks. 

We utilized Token Gradient Regularization (TGR), which has demonstrated effective transferability to CNNs, to illustrate that active vision methods exhibit a level of robustness against these attacks compared to passive baseline methods.

%% file: Tables_App/Main_Table_DenseNet121.tex
\def\arraystretch{1.2}%
\begin{table}[t!]
\centering
\caption{Accuracy (\acc) with DenseNet121 as Surrogate\vspace{-2mm}}
\label{tab:Main_densenet}
\begin{tabular}{c|c|cccc|}
\hhline{~|*5{-}|}
 & \cellcolor{Gray} & \multicolumn{1}{c|}{\cellcolor{Gray}ResNet50 \cite{He2015DeepRL}} & \multicolumn{1}{c|}{\cellcolor{Gray}Adv-T$\star$ \cite{madry2018towards} } & \multicolumn{1}{c|}{\cellcolor{Gray}ResNet34 \cite{He2015DeepRL}} & \cellcolor{Gray}CutMix \cite{CutMix} \\ \hhline{~~|*4{-}|}
 
\multirow{-2}{*}{} & \multirow{-2}{*}{\cellcolor{Gray}\textbf{Clean}} & \multicolumn{1}{c|}{\cellcolor{Gray}76.15} & \multicolumn{1}{c|}{\cellcolor{Gray}47.91} & \multicolumn{1}{c|}{\cellcolor{Gray}73.30} & \cellcolor{Gray}\textbf{78.20} \\ \hhline{~|*5{-}|} \noalign{\vspace{0.35ex}} \hline

  
\multicolumn{1}{|c|}{\multirow{2}{*}{Surrogate}} & \multirow{2}{*}{Target} & \multicolumn{4}{c|}{Attack} \\ \cline{3-6}   

\multicolumn{1}{|c|}{} &  & \multicolumn{1}{c|}{PGD \cite{madry2018towards}}   & \multicolumn{1}{c|}{MIM \cite{MIM}}    & \multicolumn{1}{c|}{VMI \cite{VMI}}    & P-IFGSM \cite{GaoZhang2020PatchWise} \\ \hline

\multicolumn{1}{|c|}{\multirow{6}{*}{DenseNet121 \cite{Densenet}}}  & ResNet34  & \multicolumn{1}{c|}{33.33} & \multicolumn{1}{c|}{23.76}  & \multicolumn{1}{c|}{14.12}  & 29.22   \\ \cline{2-6}

\multicolumn{1}{|c|}{} & ResNet50 & \multicolumn{1}{c|}{30.91} & \multicolumn{1}{c|}{21.26}  & \multicolumn{1}{c|}{12.33}  & 30.45   \\ \hhline{~|*5{-}|}

\multicolumn{1}{|c|}{} & Adv-T$\star$ & \multicolumn{1}{c|}{47.63} & \multicolumn{1}{c|}{47.40}  & \multicolumn{1}{c|}{47.19}  & 46.92   \\ \hhline{~|*5{-}|}

\multicolumn{1}{|c|}{} & CutMix & \multicolumn{1}{c|}{33.25} & \multicolumn{1}{c|}{23.05}  & \multicolumn{1}{c|}{14.31}  & 35.40   \\ \hhline{~|*5{-}|}

\multicolumn{1}{|c|}{} & \cellcolor{Ours}\textbf{FALcon} & \multicolumn{1}{c|}{\cellcolor{Ours} \textbf{43.19}} & \multicolumn{1}{c|}{\cellcolor{Ours} \textbf{30.90}}  & \multicolumn{1}{c|}{\cellcolor{Ours} \textbf{20.00}}  & \cellcolor{Ours} \textbf{35.70}   \\ \hhline{~|*5{-}|}

\multicolumn{1}{|c|}{} & \cellcolor{Ours} \textbf{GFNet}  & \multicolumn{1}{c|}{\cellcolor{Ours} \textbf{55.59}} & \multicolumn{1}{c|}{\cellcolor{Ours} \textbf{46.64}}  & \multicolumn{1}{c|}{\cellcolor{Ours} \textbf{38.46}}  & \cellcolor{Ours} \textbf{45.57} \\ \hline


    

\end{tabular}
\vspace{-4mm}
\end{table}

%% file: Tables_App/TGR_multiple_pred.tex
\begin{table}[!t]
    \begin{minipage}{.7\linewidth}
    \centering
    \captionsetup{justification=centering}
\caption{Token Gradient Regularization.}
\label{tab:TGR-table}
\begin{tabular}{|c|c|ccc|}
\hline
\multirow{2}{*}{Target} & \multirow{2}{*}{Clean} & \multicolumn{3}{c|}{Surrogate}                                                             \\ \cline{3-5} 
    &   & \multicolumn{1}{c|}{ViT} & \multicolumn{1}{c|}{CaiT} & PiT \\ \hline \hline
ResNet50 \cite{He2015DeepRL}                & \textbf{93.10}                  & \multicolumn{1}{c|}{32.0}           & \multicolumn{1}{c|}{13.3}           & 10.30          \\ \hline
Adv-T$\star$ \cite{madry2018towards}                  & 28.80                  & \multicolumn{1}{c|}{27.70}          & \multicolumn{1}{c|}{26.90}          & 26.90          \\ \hline
Swin-v2 \cite{SwinV2}                & 99.00                  & \multicolumn{1}{c|}{31.10}          & \multicolumn{1}{c|}{10.3}           & 12.30          \\ \hline
RVT \cite{Mao2021TowardsRV}                    & 97.70                  & \multicolumn{1}{c|}{\textbf{34.40}}          & \multicolumn{1}{c|}{12.1}           & \textbf{14.50} \\ \hline
Adv-RVT$\star$ \cite{Mao2021TowardsRV}               & 53.40                  & \multicolumn{1}{c|}{47.80}          & \multicolumn{1}{c|}{46.30}          & 45.70          \\ \hline
Focal Net  \cite{yang2022focal}             & \textbf{99.20}         & \multicolumn{1}{c|}{\textbf{35.60}} & \multicolumn{1}{c|}{\textbf{13.20}}          & 7.00           \\ \hline
\rowcolor{Ours}
FALcon \cite{ibrayev2023exploring}   & 85.4   & \multicolumn{1}{c|}{21.00}          & \multicolumn{1}{c|}{\textbf{15.10}} & 10.50          \\ \hline
\rowcolor{Ours}
GFNet  \cite{huang2022glance}                 & 89.5                   & \multicolumn{1}{c|}{\textbf{35.00}} & \multicolumn{1}{c|}{\textbf{22.5}}  & \textbf{15.80} \\ \hline
\end{tabular}
    \end{minipage}%
    \begin{minipage}{.3\linewidth}
      \centering
      \captionsetup{justification=centering}
    \caption{Effects of Inference from Multiple Fixation Points using ResNet34 as surrogate.}
    \label{tab:multiple_pred}
    \begin{tabular}{|c|cc|}
    \hline
    \multirow{2}{*}{Target} & \multicolumn{2}{c|}{Attack} \\ \cline{2-3} 
        & \multicolumn{1}{c|}{PGD} & MIM \\ \hline \hline
    ResNet50 & \multicolumn{1}{c|}{31.46} & 20.20 \\ \hline 
   \begin{tabular}[c]{@{}c@{}}FALcon\\ (Top-1)\end{tabular} & \multicolumn{1}{c|}{49.83} & 37.01 \\ \hline
    \cellcolor{Ours} \begin{tabular}[c]{@{}c@{}}\textbf{FALcon}\\ \textbf{(Multi)}\end{tabular} & \multicolumn{1}{c|}{\cellcolor{Ours} \textbf{54.37}} & \cellcolor{Ours} \textbf{41.40} \\ \hline
    \end{tabular}
    \end{minipage} 
\end{table}

%% file: Tables_App/Any_Preds_big_table.tex
\def\arraystretch{1.2}%
\begin{table}[t!]
\centering
\caption{On Any Predictions of FALcon\vspace{-2mm}}
\label{tab:Main_Any_Preds}
\begin{tabular}{c|c|cccc|}
\hhline{~|*5{-}|}
 & \cellcolor{Gray} & \multicolumn{1}{c|}{\cellcolor{Gray}ResNet50 \cite{He2015DeepRL}} & \multicolumn{1}{c|}{\cellcolor{Gray}Adv-T$\star$ \cite{madry2018towards} } & \multicolumn{1}{c|}{\cellcolor{Gray}\textbf{FALcon\cite{ibrayev2023exploring}}} & \cellcolor{Gray}\textbf{GFNet \cite{huang2022glance}} \\ \hhline{~~|*4{-}|}
 
\multirow{-2}{*}{} & \multirow{-2}{*}{\cellcolor{Gray}\textbf{Clean}} & \multicolumn{1}{c|}{\cellcolor{Gray}76.15} & \multicolumn{1}{c|}{\cellcolor{Gray}47.91} & \multicolumn{1}{c|}{\cellcolor{Gray}\textbf{72.97}} & \cellcolor{Gray}\textbf{75.88} \\ \hhline{~|*5{-}|} \noalign{\vspace{0.35ex}} \hline

  
\multicolumn{1}{|c|}{\multirow{2}{*}{Surrogate}} & \multirow{2}{*}{Target} & \multicolumn{4}{c|}{Attack} \\ \cline{3-6}   

\multicolumn{1}{|c|}{} &  & \multicolumn{1}{c|}{PGD \cite{madry2018towards}}   & \multicolumn{1}{c|}{MIM \cite{MIM}}    & \multicolumn{1}{c|}{VMI \cite{VMI}}    & P-IFGSM \cite{GaoZhang2020PatchWise} \\ \hline

\multicolumn{1}{|c|}{\multirow{5}{*}{ResNet34 \cite{He2015DeepRL}}}  & ResNet50  & \multicolumn{1}{c|}{31.46} & \multicolumn{1}{c|}{20.20}  & \multicolumn{1}{c|}{11.61}  & 31.62   \\ \cline{2-6} 

\multicolumn{1}{|c|}{} & Adv-T$\star$ & \multicolumn{1}{c|}{47.63} & \multicolumn{1}{c|}{47.37}  & \multicolumn{1}{c|}{47.16}  & 46.94   \\ \hhline{~|*5{-}|}

\multicolumn{1}{|c|}{} & \cellcolor{Ours}\textbf{FALcon-Top} & \multicolumn{1}{c|}{\cellcolor{Ours} \textbf{49.83}} & \multicolumn{1}{c|}{\cellcolor{Ours} \textbf{37.01}}  & \multicolumn{1}{c|}{\cellcolor{Ours} \textbf{29.40}}  & \cellcolor{Ours} \textbf{40.46}   \\ \hhline{~|*5{-}|}

\multicolumn{1}{|c|}{} & \cellcolor{Ours}\textbf{FALcon-Any} & \multicolumn{1}{c|}{\cellcolor{Ours} \textbf{53.76}} & \multicolumn{1}{c|}{\cellcolor{Ours} \textbf{41.28}}  & \multicolumn{1}{c|}{\cellcolor{Ours} \textbf{33.74}}  & \cellcolor{Ours} \textbf{44.72}   \\ \hhline{~|*5{-}|}

\multicolumn{1}{|c|}{} & \cellcolor{Ours} \textbf{GFNet}  & \multicolumn{1}{c|}{\cellcolor{Ours} \textbf{57.82}} & \multicolumn{1}{c|}{\cellcolor{Ours} \textbf{48.60}}  & \multicolumn{1}{c|}{\cellcolor{Ours} \textbf{41.17}}  & \cellcolor{Ours} \textbf{46.93} \\ \hline \hline

    
\multicolumn{1}{|c|}{\multirow{5}{*}{ResNet50 \cite{He2015DeepRL}}}  & ResNet50               & \multicolumn{1}{c|}{0.00}  & \multicolumn{1}{c|}{0.00}   & \multicolumn{1}{c|}{0.00}   & 0.00    \\ \hhline{~|*5{-}|}
    
\multicolumn{1}{|c|}{} & Adv-T$\star$  & \multicolumn{1}{c|}{47.68} & \multicolumn{1}{c|}{47.36}  & \multicolumn{1}{c|}{47.22}  & 46.83   \\ \hhline{~|*5{-}|}

\multicolumn{1}{|c|}{} & \cellcolor{Ours} \textbf{FALcon-Top}  & \multicolumn{1}{c|}{\cellcolor{Ours} \textbf{31.73}} & \multicolumn{1}{c|}{\cellcolor{Ours} \textbf{19.96}}  & \multicolumn{1}{c|}{\cellcolor{Ours} \textbf{12.07}}  &   \cellcolor{Ours} \textbf{27.37}      \\ \hhline{~|*5{-}|}

\multicolumn{1}{|c|}{} & \cellcolor{Ours} \textbf{FALcon-Any}  & \multicolumn{1}{c|}{\cellcolor{Ours} \textbf{37.54}} & \multicolumn{1}{c|}{\cellcolor{Ours} \textbf{25.60}}  & \multicolumn{1}{c|}{\cellcolor{Ours} \textbf{16.64}}  &   \cellcolor{Ours} \textbf{32.37}      \\ \hhline{~|*5{-}|}
    
\multicolumn{1}{|c|}{} & \cellcolor{Ours} \textbf{GFNet} & \multicolumn{1}{c|}{\cellcolor{Ours} \textbf{51.85}} & \multicolumn{1}{c|}{\cellcolor{Ours} \textbf{42.16}}  & \multicolumn{1}{c|}{\cellcolor{Ours} \textbf{32.33}}  & \cellcolor{Ours} \textbf{43.33}   \\ \hline
    

\end{tabular}
\vspace{-2mm}
\end{table}